\documentclass[10pt, conference, compsocconf]{IEEEtran}
\IEEEoverridecommandlockouts
\usepackage{times}
\usepackage{helvet}
\usepackage{courier}
\usepackage[hyphens]{url}
\usepackage{graphicx}
\usepackage[font=scriptsize]{subcaption}
\usepackage{amssymb,amsmath}
\usepackage[font=scriptsize]{caption}
\usepackage{multirow}
\usepackage[noend]{algpseudocode}
\usepackage{algorithm}
\algrenewcommand\alglinenumber[1]{\scriptsize #1:}
\usepackage[short]{optidef}
\usepackage{amsthm,epsfig}	
\newtheorem{theorem}{Theorem}
\newtheorem{lem}{Lemma}

\newtheorem{assumption}{Assumption}
\newtheorem{corollary}{Corollary}

\usepackage{cite,amsthm,epsfig}								
\usepackage[shortlabels]{enumitem}
\usepackage{float}
\usepackage{stfloats}
\usepackage{xcolor, array, stfloats}
\usepackage{indentfirst}
\usepackage{booktabs}

\newcommand{\rom}[1]{\uppercase\expandafter{\romannumeral #1}}

\title{Real-Time Edge Intelligence in the Making: A Collaborative Learning Framework via Federated Meta-Learning}
\author{\IEEEauthorblockN{Sen Lin, Guang Yang and Junshan Zhang\thanks{This work was supported in part by NSF under Grant CPS-1739344, ARO under grant W911NF-16-1-0448, and the DTRA under Grant HDTRA1-13-1-0029.}}
\IEEEauthorblockA{School of EECE, Arizona State University, Tempe, AZ 85287\\
\{slin70, gyang57, junshan.zhang\}@asu.edu}}
\date{}							

\begin{document}
\maketitle

\begin{abstract}
Many IoT applications at the network edge demand intelligent decisions in a real-time manner.
The edge device alone, however, often cannot achieve real-time edge intelligence due to its constrained computing resources and limited  local data. 
To tackle these challenges, we propose a platform-aided collaborative learning framework where a model is first trained across a set of source edge nodes by a federated meta-learning approach, and  then it is rapidly adapted to learn a new task at the target edge node, using a few samples only.
Further, we investigate the convergence of the proposed federated meta-learning algorithm under mild conditions on node similarity and the adaptation performance at the target edge. 
To combat against the vulnerability of meta-learning algorithms to possible adversarial attacks, we further propose a robust version of the federated meta-learning algorithm based on distributionally robust optimization, and establish its convergence under mild conditions. Experiments on different datasets demonstrate the effectiveness of the proposed Federated Meta-Learning based framework.
\end{abstract}

\section{Introduction}
 
Since most of IoT devices reside  at the network edge, pushing the AI frontier to achieve real-time edge intelligence is highly nontrivial, due to the requirements in performance, cost and privacy. Clearly, the conventional wisdom of transporting
the data bulks from the IoT devices to the cloud datacenters for analytics would not work well, simply because the  requirements  in  terms  of  high  bandwidth  and low  latency    would be  extremely  demanding  and  stringent.  As a result, it is anticipated that a high percentage of IoT data will be stored and processed locally.  However,
running AI applications directly  on edge devices to process the IoT data locally, if not designed intelligently, would suffer from
poor performance and energy inefficiency, simply because many AI applications typically require high computational power that greatly
outweighs the capacity of resource- and energy-constrained IoT devices.
To address the above challenges, edge computing has recently emerged, and the marriage of edge computing and AI has given rise to a new research area, namely ‘edge intelligence’ or ‘edge AI’ \cite{wang2019edge}\cite{ZhouCLZLZ19}.


It is highly challenging for a single edge node alone to achieve real-time edge intelligence. 
A key observation is that learning tasks across edge nodes  often share some similarity,  which can be leveraged to tackle these challenges. With this insight, we propose a platform-aided collaborative learning framework where the model knowledge is learnt collaboratively by a federation of edge nodes, in a distributed manner, and then is transferred via the platform to the target edge node for fine-tuning with its local dataset. Then, the next key question to ask is {\em ``What knowledge should the federation of  edge nodes learn and be transferred to the target edge node for achieving real-time edge intelligence?"}  


Federated learning has recently been developed for model training across multiple edge nodes, where a single global model is trained across all   edge nodes in a distributed manner. It has been shown that limited performance is achieved when fine-tuning the global model for adaptation to a new (target) edge node with a small dataset \cite{ravi2016optimization}. Along a different line, federated multi-task learning \cite{smith2017federated} has been proposed to train different but related models for different nodes, aiming to deal with the model heterogeneity among edge nodes. In particular, every source edge node, which is also a target edge node, is able to learn a unique model through capitalizing the computational resource and data belonging to other nodes. This, however, inevitably requires intensive computation and communications and is time-consuming, and hence could not meet the latency requirement for real-time edge intelligence.

Building on the recent exciting advances in meta-learning \cite{schmidhuber1987evolutionary}\cite{finn2017model}, in this paper we propose a federated meta-learning approach to address the questions mentioned above. 
The underlying rationale behind meta-learning is to train the model's
initial parameters over many tasks, such that the pre-trained model can achieve maximal performance
on a new task after quick adaptation using only
a small amount of data corresponding to that new task.
Thus inspired, we advocate a federated meta-learning approach where all source edge nodes collaboratively learn a global model initialization such that maximal performance can be obtained with the model parameters updated  with only a few data samples at the target edge node,  thereby achieving real-time edge intelligence.

Different from meta-learning which requires the knowledge of the task distribution \cite{finn2017model}, the federated meta-learning proposed in this work removes this assumption, by making use of the fact that different edge nodes often have distinct local models while sharing some similarity, and it can automate the process of task construction because each edge node is continuously making intelligent decisions. 
Further, the federated meta-learning  eliminates the need of centralized computation and thus offers the flexibility to strike a good balance between the communication cost and local computation cost (e.g., via controlling the number of local update steps).

The main contributions in this paper can be summarized as follows.
\begin{itemize}
    \item We propose a platform-aided collaborative learning framework where a model is first trained by a federated meta-learning approach across multiple  edge nodes, and then is transferred via the platform to the target edge node such that rapid adaptation can be achieved with the model updated (e.g., through one gradient step) with small local datasets, in order to achieve real-time edge intelligence. To the best of our knowledge, this is the first work to apply meta-learning to obtain real-time edge intelligence in a distributed manner.
    
    \item We  study the convergence behavior of the federated meta-learning algorithm and examine the adaptation performance of the fine-tuned model at the target edge node. In particular, we are the first to investigate the impact of the node similarity and the number of local update steps on 
    the convergence subject to communication and computation cost constraints. To establish the convergence,  we impose bounds on the variations of the gradients and Hessians of local loss functions (with respect to the hyper-parameter) across edge nodes, thereby removing the assumption in meta-learning that requires all tasks follow a (known) distribution.
    
    \item To combat against the possible vulnerability of meta-learning algorithms, we propose a robust version of federated meta-learning, building on recent advances in distributionally robust optimization (DRO). We further show that the proposed algorithm still converges under mild technical conditions. To the best of our knowledge, this is the first work to exploit DRO for improving the robustness of meta-learning algorithms. 
    \item We evaluate the performance of the proposed collaborative learning framework using different datasets, which corroborates the effectiveness of federated meta-learning and showcases the robustness of the DRO-based robust federated meta-learning.
\end{itemize}

\section{Related Work}
The concept of meta-learning is not new, but the recent advances through gradient-based optimization bring it into the light again as a promising solution for fast learning. 
In particular, Finn et al. \cite{finn2017model} propose one gradient-based algorithm called MAML, which directly optimizes the learning performance with respect to an initialization of the model such that even one-step gradient descent from that initialization can still produce good results on a new task.
To circumvent the need of the second derivatives in MAML, Nichol et al. \cite{nichol2018first} propose a first-order method named Reptile, which is similar to joint training but surprisingly works well as a meta-learning algorithm. 
A similar meta-learning framework is studied by Chen et al. \cite{chen2018} for recommendation systems through assigning one task to every user, which nevertheless does not consider any system modeling in federated learning.

To the best of our knowledge, we are among the first to establish the convergence of (federated) meta-learning. During the preparation of  this work, the preprint of one concurrent work \cite{fallah2019convergence} about convergence analysis of meta-learning algorithms became available online. It is worth noting that \cite{fallah2019convergence} studies the convergence of centralized MAML algorithms for non-convex functions, whereas this paper focuses on the convergence and adaptation performance of the federated meta-learning algorithm with  node similarity to achieve real-time edge intelligence in a federated setting, where multiple local update steps are allowed to balance the trade-off between the communication cost and local computation cost. 

The susceptibility of meta-learning algorithms such as MAML to adversarial attacks is first investigated in \cite{edmundstransferability}. And recently \cite{yin2018adversarial} also demonstrates the significant performance degradation of MAML with adversarial samples. To make meta-learning more robust, \cite{yin2018adversarial} proposes a meta-learning algorithm called ADML which exploits both clean and adversarial samples to push the inner gradient update to arm-wrestle with the meta-update. Unfortunately, this type of approaches are generally intractable. The DRO-based robust federated meta-learning algorithm proposed in this work is not only computationally tractable, but also resistant to more general perturbations, e.g., out-of-distribution samples. In addition, the trade-off between robustness and accuracy can be fine-tuned by the size of the distributional uncertainty set.

Federated learning is first proposed by Mcmahan et al. \cite{mcmahan2016communication} which performs a variable number of local updates on a subset of devices to enable flexible and efficient communication patterns but without any theoretical convergence guarantee. Based on this, Wang et al. \cite{wang2019} analyze the convergence  with fixed number of local updates for non-independent and identically distributed (i.i.d) data distributions across devices, and introduce a control algorithm to dynamically adapt the frequency of global aggregation to minimize the loss under fixed resource constraints. Departing from the need of manually tuning the number of local updates, Sahu et al. \cite{sahu2018convergence} propose a more generalized algorithm, so called FedProx, to tackle the statistical heterogeneity inherent in federated learning and characterize the convergence behaviour for non-convex loss functions. Notwithstanding, federated learning is not designed for fast learning with small datasets. In particular, federated learning intends to find a global model that fits the data as accurately as possible for all participating nodes, wheras federated meta-learning learns a model initialization, from which fast adaptation from even small datasets can still reach good performance, and also keeps the node heterogeneity in the sense that different models would be learnt for different nodes after quick adaptation from the global model initialization.

Both meta-learning and multi-task learning \cite{evgeniou2004regularized}\cite{ruder2017overview}\cite{smith2017federated} aim to improve the learning performance by leveraging other related tasks. However, meta-learning focuses on the fast learning ability with small sample sizes and the performance improvement at the target (learning at the source is irrelevant), whereas multi-task learning aims to learn both the source and target tasks simultaneously and accurately. Besides, the model initialization learned by meta-learning can be fine-tuned with good performance on various target tasks using minimal data points, while multi-task learning may favor tasks with significantly larger amount of samples than others.

\section{Federated Meta-Learning for Achieving Real-Time Edge Intelligence}

As illustrated in Figure 1, we consider a platform-aided architecture where a set $\mathcal{S}$ of source edge nodes (each with a task) join force for federated meta-learning,
and the learned model would be transferred from the platform to a target edge node $t$ (not in $\mathcal{S}$), for rapid adaptation based on its local data.  The primary objective of the proposed federated meta-learning is to train a meta-model that can quickly adapt to the task at the target edge node to achieve real-time edge intelligence, using only a few local data points. To accomplish this, the meta-model is trained during a meta-learning phase across the source edge nodes in a distributed manner.

\begin{figure}
\centering
\includegraphics[scale=0.28]{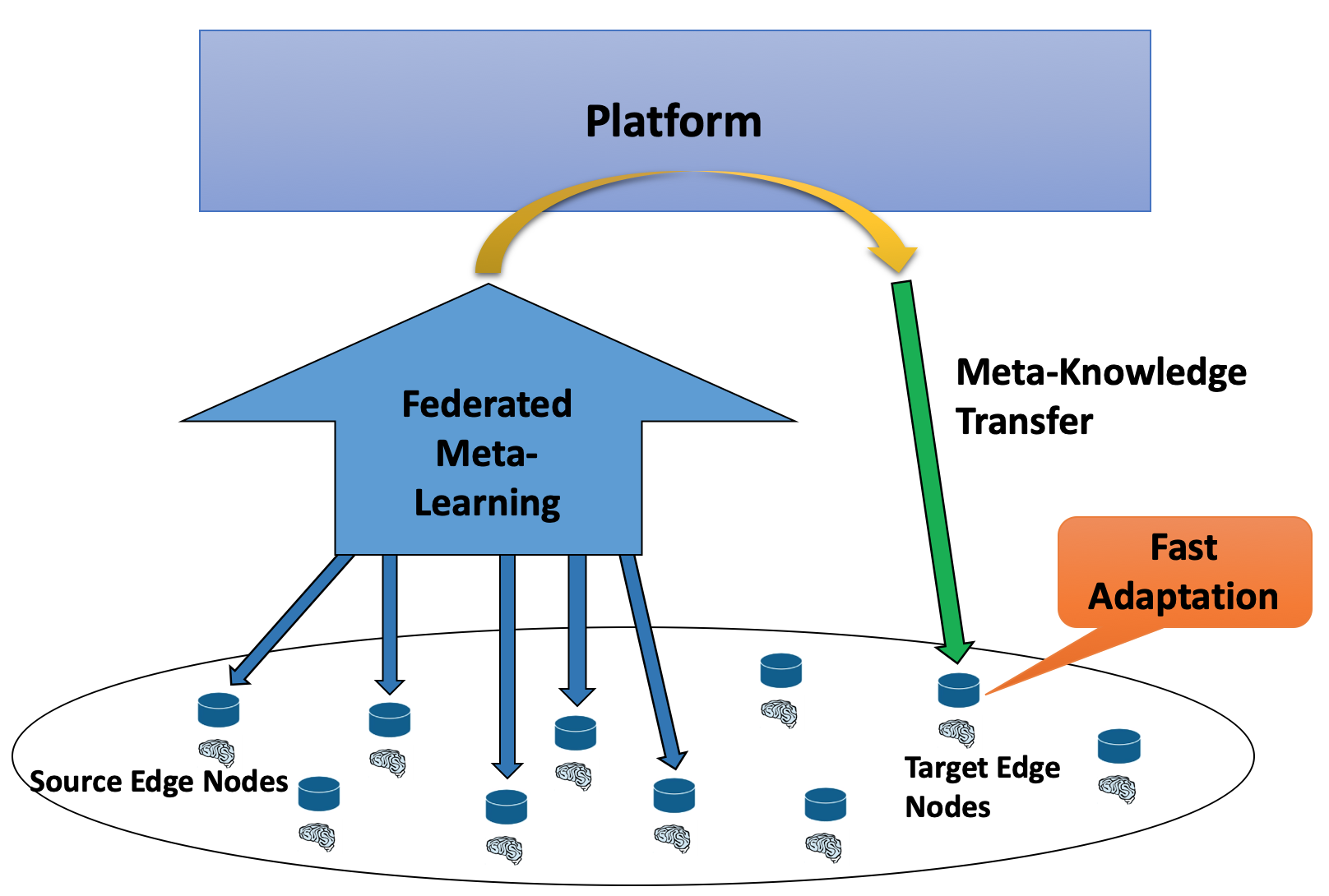}

\caption{A Platform-Aided Collaborative Learning Framework with Federated Meta-Learning for Real-Time Edge Intelligence.}
\label{Fig:system_model_1}
\vspace{-0.5cm}
\end{figure}

\subsection{Problem Formulation}
Specifically, we assume that the tasks across edge nodes follow a meta-model, represented by a parametrized function $f_{\boldsymbol{\theta}}$ with parameters $\boldsymbol{\theta}\in \mathbb{R}^d$. For source edge node $i\in\mathcal{S}$, let $D_i$ denote its local dataset $\{(\mathbf{x}_i^1, \mathbf{y}_i^1),...,(\mathbf{x}_i^j, \mathbf{y}_i^j),...,(\mathbf{x}_i^{|D_i|}, \mathbf{y}_i^{|D_i|}) \}$, where $|D_i|$ is the dataset size and $(\mathbf{x}^j, \mathbf{y}^j)\in \mathcal{X}\times \mathcal{Y}$ is a sample point with  $\mathbf{x}^j$ being the input and $\mathbf{y}^j$  the output. 
We further assume that $(\mathbf{x}_i^j, \mathbf{y}_i^j)$ follows an unknown distribution $P_i$.
Denote the loss function by $l(\boldsymbol{\theta}, (\mathbf{x}^j, \mathbf{y}^j)):\mathcal{X}\times \mathcal{Y}\rightarrow \mathbb{R}$.  The empirical loss function for node $i$ is then defined as
\vspace{-0.1cm}
\begin{equation} \label{edgeloss}
    L(\boldsymbol{\theta}, D_i)\triangleq \frac{1}{|D_i|}\sum_{(\mathbf{x}_i^j,\mathbf{y}_i^j)\in D_i} l(\boldsymbol{\theta},(\mathbf{x}_i^j,\mathbf{y}_i^j)),
    \vspace{-0.1cm}
\end{equation}
which we write as $L_i(\boldsymbol{\theta})$ for brevity.  
Moreover, we use $L_w(\boldsymbol{\theta})$ to denote the overall loss function across all edge nodes in $\mathcal{S}$:
\vspace{-0.2cm}
\begin{equation} \label{globalloss}
    L_w(\boldsymbol{\theta})\triangleq \sum_{i\in\mathcal{S}}\omega_i L_i(\boldsymbol{\theta}),
    \vspace{-0.2cm}
\end{equation}    
where $\omega_i=\frac{|D_i|}{\sum_{i\in\mathcal{S}}|D_i|}$ and
the weight $\omega_i$ of each edge node depends on its own local data size. 

In the same spirit as MAML, we consider that the target edge node $t$ has $K$ data samples, i.e., $|D_t|=K$. For each source edge node $i\in\mathcal{S}$, $D_i$ is divided into two disjoint sets, the training set $D_i^{train}$ and the testing set $D_i^{test}$, where $|D_i^{train}|=K$ (assuming $|D_i|>K$ for all $i\in\mathcal{S}$). Given the model parameter $\boldsymbol{\theta}$, the edge node $i$ first updates $\boldsymbol{\theta}$ using one step gradient descent based on $D_i^{train}$:
\vspace{-0.1cm}
\begin{equation} \label{gradientupdate}
    \boldsymbol{\phi}_i(\boldsymbol{\theta})=\boldsymbol{\theta}-\alpha \nabla_{\boldsymbol{\theta}} L(\boldsymbol{\theta}, D_i^{train}),
    \vspace{-0.1cm}
\end{equation}
with $\alpha$ being the learning rate, and then evaluates the loss $L(\boldsymbol{\phi}_i,D_i^{test})$ for the updated model parameter $\boldsymbol{\phi}_i$ based on $D_i^{test}$. It follows that the overall objective of the federated meta-learning is given by 
\begin{mini}
	{\boldsymbol{\theta}}{\sum_{i\in\mathcal{S}}\omega_i L(\boldsymbol{\phi}_i(\boldsymbol{\theta}),D_i^{test}).}
	{\label{cloudmeta}}{}
\end{mini}
 Intuitively, by considering how the test error on local testing datasets changes with respect to the updated model parameters, we aim to obtain a model initialization such that small changes in the model parameters, i.e., altering in the direction of the loss gradient, would lead to substantial performance improvements for any task across the edge nodes.

Departing from  MAML which assumes that a distribution over tasks is given, we do not require such an assumption here. Instead, we will quantify the node similarity in terms of the variations of the gradients and Hessians of local loss functions with respect to the hyper-parameter.
Worth noting is that without knowing the data, the platform cannot directly solve the problem \eqref{cloudmeta}.

\subsection{Federated Meta-Learning (FedML)}
Motivated by federated learning, we propose to solve the meta-learning objective \eqref{cloudmeta} in a distributed manner, in the sense that each node locally updates the model parameter $\boldsymbol{\theta}$ based on its own dataset and transmits the updated value to the platform for a global aggregation. To better utilize the local computing resource of edge nodes and reduce the communication cost between the platform and edge nodes which is often a significant bottleneck in wireless networks, each edge node is allowed to locally update $\boldsymbol{\theta}$ for $T_0$ steps before uploading the results to the platform. 

\textbf{Federated Meta-Training across Source Nodes:} More specifically, the platform transfers an initialized $\boldsymbol{\theta}^0$ to all nodes in $\mathcal{S}$ at time $t=0$. 
As outlined in Algorithm 1, there are two major steps: 
\begin{itemize}
\item\textbf{Local Update}: For $t\neq nT_0$ where $n\in\mathbb{N}^{+}$, each node $i\in\mathcal{S}$ first updates $\boldsymbol{\theta}_i^t$ using the training dataset $D_i^{train}$ based on \eqref{gradientupdate} and then locally updates $\boldsymbol{\theta}_i^t$ again through evaluating $\boldsymbol{\phi}_i^t$ on the testing dataset $D_i^{test}$:
\begin{equation} \label{testupdate}
    \boldsymbol{\theta}_i^{t+1}=\boldsymbol{\theta}_i^t-\beta\nabla_{\boldsymbol{\theta}} L(\boldsymbol{\phi}_i^t, D_i^{test}),
\end{equation}
where $\beta$ is the meta learning rate and $\boldsymbol{\theta}_i^{t+1}$ will be used as the starting point for the next iteration at node $i$. 
\item\textbf{Global Aggregation:} For $t=nT_0$, each node also needs to transmit the updated $\boldsymbol{\theta}_i^{t+1}$ to the platform. The platform then performs a global aggregation to achieve $\boldsymbol{\theta}^{t+1}$:
\begin{equation} \label{aggregation}
    \boldsymbol{\theta}^{t+1}=\sum_{i\in\mathcal{S}}\omega_i\boldsymbol{\theta}_i^{t+1}, 
\end{equation}
and sends $\boldsymbol{\theta}^{t+1}$ back to all edge nodes for the next iteration.
\end{itemize}
The details are summarized in Algorithm 1.
\begin{algorithm}[H]
  \footnotesize
	\caption{Federated meta-learning (FedML)}
	\label{alg1}
 	\begin{algorithmic}[1]
		  \Statex\textbf{Inputs:} $M$, $K$, $T$, $T_0$, $\alpha$, $\beta$, $\omega_i$ for $i\in\mathcal{S}$\;
		  \Statex\textbf{Outputs:} Final model parameter $\boldsymbol{\theta}$\;
		  \State Platform randomly initializes $\boldsymbol{\theta}^0$ and sends it to all edge nodes in $\mathcal{S}$;
		  \For{$t= 1, 2, ..., T$}
		    \For{each node $i\in\mathcal{S}$}
		          \State Compute the updated parameter with one-step gradient descent using $D_i^{train}$: $\boldsymbol{\phi}_i^{t}=\boldsymbol{\theta}_i^t-\alpha \nabla_{\boldsymbol{\theta}}L(\boldsymbol{\theta}_i^t, D_i^{train})$;
		         \State Obtain $\boldsymbol{\theta}_i^{t+1}$ based on \eqref{testupdate} using $D_i^{test}$;
		        //\textbf{local update}
		        \If{$t$ is a multiple of $T_0$}
		          \State Send $\boldsymbol{\theta}_i^{t+1}$ to the platform;
		          \State Receive $\boldsymbol{\theta}^{t+1}$ from the platform where $\boldsymbol{\theta}^{t+1}$ is obtained based on \eqref{aggregation};
		          \State Set $\boldsymbol{\theta}_i^{t}\leftarrow \boldsymbol{\theta}^{t+1}$; //\textbf{global aggregation}
		         \Else
		         \State Set $\boldsymbol{\theta}_i^{t}\leftarrow \boldsymbol{\theta}_i^{t+1}$;
		      \EndIf
		    \EndFor
		\EndFor
		\State\textbf{return} $\boldsymbol{\theta}$.
	\end{algorithmic}
\end{algorithm}

\textbf{Fast Adaptation towards Real-time Edge Intelligence at Target Node:} Given the model parameter $\boldsymbol{\theta}$ from the platform, the target edge node $t$ can quickly adapt the model based on its local dataset $D_t$ and obtain a new model parameter $\boldsymbol{\phi}_t$ through one step gradient descent:
\begin{equation} \label{fastadapt}
    \boldsymbol{\phi_t}=\boldsymbol{\theta}-\alpha\nabla_{\boldsymbol{\theta}} L(\boldsymbol{\theta}, D_t).
\end{equation}

In a nutshell, instead of training each source edge node to learn a global model as in federated learning, the source nodes join force to learn how to learn quickly with only a few data samples in a distributed manner, i.e., learn $\boldsymbol{\theta}$ such that just one step gradient descent from $\boldsymbol{\theta}$ can bring up a new model suitable for a specific target node. This greatly improves the fast learning capability at the edge with the collaboration among edge nodes.

\section{Performance Analysis}

 In this section, we seek to answer the following two key questions : (1) What is the convergence performance of the proposed federated meta-learning algorithm? (2) Can the fast adaptation at the target node achieve good performance?  

\subsection{Convergence Analysis}
For ease of exposition, we define function $G_i(\boldsymbol{\theta})\triangleq L_i(\boldsymbol{\phi}_i(\boldsymbol{\theta}))$ and $G(\boldsymbol{\theta})\triangleq\sum_{i\in\mathcal{S}}\omega_i G_i(\boldsymbol{\theta})$ such that problem \eqref{cloudmeta} can be written as:
\begin{mini}
	{\boldsymbol{\theta}}{G(\boldsymbol{\theta}).}
	{\label{convergence}}{}
\end{mini}
For convenience, we assume $T=NT_0$ and make the following assumptions to the loss function for all $i\in\mathcal{S}$:
\begin{assumption}
Each $L_i(\boldsymbol{\theta})$ is $\mu$-strongly convex, i.e., for all $\boldsymbol{\theta}$, $\boldsymbol{\theta}'\in \mathbb{R}^d$,
    \begin{equation*}
        \langle \nabla L_i(\boldsymbol{\theta})-\nabla L_i(\boldsymbol{\theta}'), \boldsymbol{\theta}-\boldsymbol{\theta}'\rangle\geq \mu \|\boldsymbol{\theta}-\boldsymbol{\theta}'\|^2.
    \end{equation*}
\end{assumption}
\begin{assumption}
Each $L_i(\boldsymbol{\theta})$ is $H$-smooth, i.e., for all $\boldsymbol{\theta}$, $\boldsymbol{\theta}'\in \mathbb{R}^d$,
    \begin{equation*}
        \|\nabla L_i(\boldsymbol{\theta})-\nabla L_i(\boldsymbol{\theta}')\|\leq H\|\boldsymbol{\theta}-\boldsymbol{\theta}'\|,
    \end{equation*}
and there exist constant $B$ such that for all $\boldsymbol{\theta}\in\mathbb{R}^d$
    \begin{equation*}
       \|\nabla L_i(\boldsymbol{\theta})\|\leq B.
    \end{equation*}
\end{assumption}
\begin{assumption}
The Hessian of each $L_i(\boldsymbol{\theta})$ is $\rho$-Lipschitz, i.e., for all $\boldsymbol{\theta}$, $\boldsymbol{\theta}'\in \mathbb{R}^d$,
    \begin{equation*}
        \|\nabla^2 L_i(\boldsymbol{\theta})-\nabla^2 L_i(\boldsymbol{\theta}')\|\leq \rho\|\boldsymbol{\theta}-\boldsymbol{\theta}'\|.
    \end{equation*}
\end{assumption}
\begin{assumption}
There exists constants $\delta_i$ and $\sigma_i$ such that for all $\boldsymbol{\theta}\in\mathbb{R}^d$ 
    \begin{align*}
        \|\nabla L_i(\boldsymbol{\theta})-\nabla L_w(\boldsymbol{\theta})\|\leq \delta_i, \\
        \|\nabla^2 L_i(\boldsymbol{\theta})-\nabla^2 L_w(\boldsymbol{\theta})\|\leq \sigma_i.
    \end{align*}
\end{assumption}
Assumptions 1-2 are standard and hold in many machine learning applications, e.g., in logistic regression over a bounded domain and squared-SVM. Assumption 3 is concerned with the high-order smoothness of the local loss function at each edge node, which makes it possible to characterize the landscape of the local meta-learning objective function. Assumption 4 is imposed to capture the node similarity. Specifically, we impose the condition that the variations of the gradients and Hessians of local loss functions (with respect to the hyper-parameter) across edge nodes are upper bounded by some constant. 
Intuitively, a small (large) constant indicates that the tasks are more (less) similar, and this constant can be tuned over a wide range to obtain a general understanding.  
Further, for a task distribution with task gradients uniformly bounded above (as is standard), Assumption 4 follows directly because $\|\nabla L_i(\boldsymbol{\theta})-\nabla L_w(\boldsymbol{\theta})\|$ (also Hessian) can be viewed as the distance between a typical realization and the sample average. In a nutshell, the task similarity assumption here is more general and realistic than the assumption in meta-learning that all tasks follow a (known) distribution.
It is worth noting that these assumptions do not trivialize the meta-learning setting.

To characterize the convergence behavior of the federated meta-learning algorithm, we first examine the structural properties of the global meta-learning objective $\mathbf{G}(\boldsymbol{\theta})$. Next, we study the impact of the task similarity across different edge nodes on the convergence performance of the federated meta-learning, which is further complicated by multiple local updates at each node to reduce the communication overhead. 

\indent\textit{Convexity and Smoothness of the Federated Meta-Learning Objective Function:}  Based on Theorem 1 in \cite{finn2019online}, we first have the following result about the structural properties of function $G(\boldsymbol{\theta})$.
\begin{lem} 
Suppose Assumptions 1-3 hold. When $\alpha\leq \min\{\frac{\mu}{2\mu H+\rho B},\frac{1}{\mu}\}$, $G(\boldsymbol{\theta})$ is $\mu'$-strongly convex and $H'$-smooth, where $\mu'=\mu(1-\alpha H)^2-\alpha\rho B$ and $H'=H(1-\alpha\mu)^2+\alpha\rho B$.
\end{lem}
 
Lemma 1 indicates that when the learning rate $\alpha$ is relatively small, the meta-learning objective function $G(\boldsymbol{\theta})$ formed by the one-step gradient descent on local datasets is as well-behaved as the local loss function.

\indent\textit{Bounded Dissimilarity across Local Learning tasks:}
Next, we characterizes the impact of the similarity across   local  learning tasks.
\begin{theorem}
Suppose Assumptions 2 and 4 hold. Then there exists a constant $C$ such that
    \begin{equation*}
        \|\nabla G_i(\boldsymbol{\theta})-\nabla G(\boldsymbol{\theta})\|\leq \delta_i+\alpha C(H\delta_i+B\sigma_i+\tau),
    \end{equation*}
    where $\tau=\sum_{i\in\mathcal{S}} \omega_i\delta_i\sigma_i$.
\end{theorem}
Given the bounded variance of gradients and Hessians of local loss functions, we can  find upper bounds on the gradient variance of the local meta-learning objective functions, while still preserving the node heterogeneity.
As a sanity check, if all the edge nodes have same data points, it follows that $\delta_i=\sigma_i=0$ for all $i\in\mathcal{S}$. Consequently, all edge nodes have the same local learning objective. 
 

Based on Lemma 1 and Theorem 1, we can have the following result about the convergence performance of the federated meta-learning algorithm.
\begin{theorem}
Suppose that Assumptions 1-4 hold, and the learning rates $\alpha$ and $\beta$ are chosen to satisfy that $\alpha\leq\min\{\frac{\mu}{2\mu H+\rho B},\frac{1}{\mu}\}$ and $\beta<\min\{\frac{1}{2\mu'},\frac{2}{H'}\}$. Let $\delta=\sum_{i\in\mathcal{S}}\omega_i\delta_i$ and $\sigma=\sum_{i\in\mathcal{S}}\omega_i\sigma_i$. Then
    \begin{equation*}
        G(\boldsymbol{\theta}^T)-G(\boldsymbol{\theta}^\star)\leq \xi^T[G(\boldsymbol{\theta}^0)-G(\boldsymbol{\theta}^\star)]+\frac{B(1-\alpha\mu)}{1-\xi^{T_0}}h(T_0),
    \end{equation*}
    where 
    $\xi=1-2\beta\mu'\left(1-\frac{H'\beta}{2}\right)$, $h(x)\triangleq\frac{\alpha'}{\beta H'}[(1+\beta H')^{x}-1]-\alpha'x$, $\alpha'=\beta[\delta+\alpha C(H\delta+B\sigma+\tau)]$.
\end{theorem}

Intuitively, the term $\frac{B(1-\alpha\mu)}{1-\xi^{T_0}}h(T_0)$ captures the error introduced by both task dissimilarity and multiple local updates through the function $h(T_0)$. Specifically, observe that $h(T_0)$ increases with $\delta$ and $\sigma$, which clearly indicates how the task similarity and the number of local update steps $T_0$ impact the convergence performance, i.e., given a fixed duration $T$ the convergence error decreases with the task similarity while increasing with the number of local update steps when $T_0$ is large. Correspondingly, the platform is able to balance between the platform-edge communication cost and the local computation cost via controlling the number of local update steps $T_0$ per communication round, depending on the task similarity across the edge nodes.

Different from MAML, multiple local updates are allowed in Algorithm 1 to reduce the communication cost, which has nontrivial impact on the convergence behavior. As shown in Theorem 2, the convergence gap for federated meta-learning would be  large if the number of local update steps $T_0$ is large even if the tasks are very similar.
When $T_0=1$, i.e., each edge node only updates the model locally for one iteration, the term $\frac{B(1-\alpha\mu)}{1-\xi^{T_0}}h(T_0)$ disappears because $h(1)=0$.
We have the following result for this case.
\begin{corollary}
Suppose that Assumptions 1-4 hold, and the learning rates $\alpha$ and $\beta$ are chosen to satisfy that $\alpha\leq\min\{\frac{\mu}{2\mu H+\rho B},\frac{1}{\mu}\}$ and $\beta<\min\{\frac{1}{2\mu'},\frac{2}{H'}\}$. When $T_0=1$, $G(\boldsymbol{\theta}^T)-G(\boldsymbol{\theta}^\star)\leq\xi^T[G(\boldsymbol{\theta}^0)-G(\boldsymbol{\theta}^\star)]$.
\end{corollary}

\subsection{Performance Evaluation of Fast Adaptation}

The fast learning performance at the target edge node $t$ depends on not only its local sample size $D_t$ but also the similarity with the source edge nodes in the federated meta-learning. Denote $\boldsymbol{\theta}_c$ as the output of the federated meta-learning at the platform and $\boldsymbol{\theta}_c^\star$ as the optimal meta-learning model.
We assume that the convergence error $\|\boldsymbol{\theta}_c-\boldsymbol{\theta}_c^\star\|$ of the federated meta-learning algorithm is upper bounded by $\epsilon_c$.
For convenience, we further define $L^\star_t(\boldsymbol{\theta})$ as the local average loss over the underlying data distribution $P_t$:
\begin{equation} \label{trueloss}
    L^\star_t(\boldsymbol{\theta})\triangleq \mathbb{E}_{(\mathbf{x}_t^j, \mathbf{y}_t^j)\sim P_t} l(\boldsymbol{\theta},(\mathbf{x}_t^j,\mathbf{y}_t^j)).
\end{equation}
Then, the empirical loss $L_t(\boldsymbol{\theta})$ is the sample average approximation of $L^\star_t(\boldsymbol{\theta})$. Let $\boldsymbol{\phi}_t=\boldsymbol{\theta}_c-\alpha\nabla L_t(\boldsymbol{\theta}_c)$ and $\boldsymbol{\phi}^\star_t=\arg\min L^\star_t(\boldsymbol{\phi})=\boldsymbol{\theta}_t^\star-\alpha\nabla L^\star_t(\boldsymbol{\theta}_t^\star)$.
The following result characterizes the trade-off between the target-source similarity and local sample size.
\begin{theorem}
Suppose $l(\boldsymbol{\theta},(\mathbf{x}_t^j, \mathbf{y}_t^j))$ is $H$-smooth with respect to $\boldsymbol{\theta}$ for all $\boldsymbol{\theta}\in\mathbb{R}^d$ and $(\mathbf{x}_t^j, \mathbf{y}_t^j)\sim P_t$. For any $\epsilon>0$, there exist positive constants $C_t$ and $\eta=\eta(\epsilon)$ such that with probability at least $1-C_te^{-K\eta}$ we can have
\begin{align*}
    \|L^\star_t(\boldsymbol{\phi}_t)-L^\star_t(\boldsymbol{\phi}^\star_t)\|
    \leq& \alpha H\epsilon+H(1+\alpha H)\epsilon_c\\
    &+H(1+\alpha H)\|\boldsymbol{\theta}_t^\star-\boldsymbol{\theta}_c^\star\|.
\end{align*}
\end{theorem}

Theorem 3 sheds light on how the task similarity and local sample size impact the learning performance at the target edge node $t$.
In particular, the performance gap between the optimal model and the model after fast adaptation is upper bounded by the surrogate difference, denoted by $\|\boldsymbol{\theta}_t^\star-\boldsymbol{\theta}_c^\star\|$, which serves as a guidance for the platform to determine how similar the source edge nodes in the federated meta-learning should be with the target node in order to achieve given learning performance via fast adaptation and hence edge intelligence at the target edge node.

\section{Robust Federated Meta-Learning (FedML)}
It has been shown in \cite{yin2018adversarial} and \cite{edmundstransferability} that meta-learning algorithms (e.g., MAML) are vulnerable to adversarial attacks, 
 leading to possible significant performance degradation of the locally fast adapted model at the target when facing perturbed data inputs. Thus motivated, we next  devise a robust federated meta-learning algorithm and study the trade-off between robustness and accuracy therein (cf.  \cite{tsipras2018robustness}).


\subsection{Robust Federated Meta-Learning}

To combat against the possible vulnerability of meta-learning  algorithms, we propose to obtain a model initialization from which the model updated with local training data at the target not only is robust against data distributions that are distance $\pi$ away from the target data distribution $P_t$, but also guarantees good performance when fed with the clean target data. 
Based on recent advances in distributionally robust optimization (DRO), this can be achieved by solving the following problem: 
\begin{equation}\label{robustobject}
    \min_{\boldsymbol{\theta}}\left\{L_t(\boldsymbol{\phi}_t)+\max_{P:D(P,P_t)\leq\pi} \mathbb{E}_P[l(\boldsymbol{\phi}_t,(\mathbf{x},\mathbf{y}))]\right\},
\end{equation}
where $D(P,P_t)$ is a distance metric on the space of probability distributions. 

To solve \eqref{robustobject} with federated meta-learning, based on the general machine learning principle that train and test conditions must match,
we can reformulate the federated meta-learning objective \eqref{cloudmeta} as:
\begin{mini}
	{\boldsymbol{\theta}}{\sum_{i\in\mathcal{S}}\omega_i F_i(\boldsymbol{\phi}_i),}
	{\label{robustmeta}}{}
	\addConstraint{}{F_i(\boldsymbol{\phi}_i)=L(\boldsymbol{\phi}_i,D_i^{test})+\max_{P} \mathbb{E}_P[l(\boldsymbol{\phi}_i,(\mathbf{x}_i,\mathbf{y}_i))]}
	\addConstraint{}{\boldsymbol{\phi}_i=\boldsymbol{\theta}-\alpha\nabla_{\boldsymbol{\theta}}L(\boldsymbol{\theta},D_i^{train})}
	\addConstraint{}{D(P,P_i)\leq\pi,}
\end{mini}
where a distributionally robust objective function similar with the target node is set for every source edge node $i\in\mathcal{S}$.

\subsection{Wasserstein Distance based Robust Federated Meta-Learning}

The choice of the distributional distance metric clearly affects both the richness of the uncertainty set and the tractability of problem \eqref{robustmeta}. To enable distance measure between distributions with different support,
we use Wasserstein distance as the distance metric on the space of probability distributions. More specifically, let the transportation cost $c:(\mathcal{X}\times \mathcal{Y})\times (\mathcal{X}\times \mathcal{Y})\rightarrow \mathbb{R}_+$ be lower-semicontinuous and satisfy $c((\mathbf{x},\mathbf{y}),(\mathbf{x},\mathbf{y}))=0$, which quantifies the cost of transporting unit mass from $(\mathbf{x},\mathbf{y})$ to $(\mathbf{x}',\mathbf{y}')$. For any two probability measure $P$ and $Q$ supported on $\mathcal{X}\times \mathcal{Y}$, let $\Pi(P,Q)$ denote the set of all couplings (transport plans) between $P$ and $Q$, meaning measures $W$ with $W(A,\mathcal{X}\times \mathcal{Y})=P(A)$ and $W(\mathcal{X}\times \mathcal{Y},A)=Q(A)$. The Wasserstein distance is then defined as 
\begin{equation}\label{wass}
    D_w(P,Q)\triangleq \inf_{W\in\Pi(P,Q)}\mathbb{E}_W[c((\mathbf{x},\mathbf{y}),(\mathbf{x}',\mathbf{y}'))],
\end{equation}
which represents the optimal transport cost, i.e., the lowest expected transport cost, that is achievable among all couplings between $P$ and $Q$.

Since the Wasserstein distance based optimization problem is computationally demanding for complex models, based on \cite{sinha2017certifying}, we consider the following Lagrangian relaxation of the inner maximization problem of \eqref{robustmeta} with penalty parameter $\lambda\geq 0$:
\begin{equation}\label{lagrangian}
    \max_{P} \{\mathbb{E}_P[l(\boldsymbol{\phi}_i,(\mathbf{x}_i,\mathbf{y}_i))]-\lambda D_w(P,P_i)\},
\end{equation}
where $\lambda$ is inversely proportional to $\pi$.
The duality result below in \cite{blanchet2019quantifying} based on Kantorovich’s duality, a widely used approach to solve the Wasserstein distance based DRO problem in optimal transport, provides us an efficient way to solve \eqref{lagrangian} through a robust surrogate loss:
\begin{lem}
Let $l:\mathbb{R}^d\times(\mathcal{X}\times\mathcal{Y})\rightarrow \mathbb{R}$ and $c:(\mathcal{X}\times \mathcal{Y})\times (\mathcal{X}\times \mathcal{Y})\rightarrow \mathbb{R}_+$ be continuous. Define the robust surrogate loss as 
$l_{\lambda}(\boldsymbol{\theta},(\mathbf{x}_0,\mathbf{y}_0))\triangleq sup_{\mathbf{x}\in\mathcal{X}}\{l(\boldsymbol{\theta},(\mathbf{x},\mathbf{y}_0))-\lambda c((\mathbf{x},\mathbf{y}_0),(\mathbf{x}_0,\mathbf{y}_0))\}$. For any distribution $Q$ and $\lambda\geq 0$, we have
\begin{equation}
    \max_P \{\mathbb{E}_P[l(\boldsymbol{\theta},(\mathbf{x},\mathbf{y}))]-\lambda D_w(P,Q)\}=\mathbb{E}_Q[l_{\lambda}(\boldsymbol{\theta},(\mathbf{x},\mathbf{y}))].\nonumber
\end{equation}
\end{lem}
Lemma 2 divulges a worst-case joint probability measure $W^\star$ corresponding to a transport plan that transports mass from $\mathbf{x}$ to the optimizer of the local optimization problem $sup_{\mathbf{x}\in\mathcal{X}}\{l(\boldsymbol{\theta},(\mathbf{x},\mathbf{y}_0))-\lambda c((\mathbf{x},\mathbf{y}_0),(\mathbf{x}_0,\mathbf{y}_0))\}$.
Hence, we can replace \eqref{lagrangian} with the expected robust surrogate loss $\mathbb{E}_{P_i}[l_{\lambda}(\boldsymbol{\phi}_i,(\mathbf{x}_i,\mathbf{y}_i))]$. Moreover, we typically replace $P_i$ by the empirical distribution $\hat{P}_i$ because $P_i$ is unknown. 

In what follows, we focus on the following relaxed robust problem:
\begin{mini}
	{\boldsymbol{\theta}}{\sum_{i\in\mathcal{S}}\omega_i\{L(\boldsymbol{\phi}_i,D_i^{test})+\mathbb{E}_{\hat{P}_i}[l_{\lambda}(\boldsymbol{\phi}_i,(\mathbf{x}_i,\mathbf{y}_i))]\},}
	{\label{robustrelax}}{}
	\addConstraint{}{\boldsymbol{\phi}_i=\boldsymbol{\theta}-\alpha\nabla_{\boldsymbol{\theta}}L(\boldsymbol{\theta},D_i^{train}).}
\end{mini}
Under suitable conditions,  the robust surrogate loss $l_{\lambda}(\boldsymbol{\theta},(\mathbf{x}_0,\mathbf{y}_0))$ is  strongly-concave  for $\lambda\geq H_{\mathbf{x}}$ \cite{sinha2017certifying}, which indicates the computational benefits of relaxing the strict robustness requirements of \eqref{lagrangian}. Therefore,
\begin{equation}
    \nabla_{\boldsymbol{\phi}_i} l_{\lambda}(\boldsymbol{\phi}_i,(\mathbf{x}_i,\mathbf{y}_i))=\nabla_{\boldsymbol{\phi}_i} l(\boldsymbol{\phi}_i,(\mathbf{x}^\star,\mathbf{y}_i)),
\end{equation}
where
\begin{equation}\label{optimalx}
    \mathbf{x}^\star=\arg\max_{\mathbf{x}\in\mathcal{X}}\{l(\boldsymbol{\phi}_i,(\mathbf{x},\mathbf{y}_i))-\lambda c((\mathbf{x},\mathbf{y}_i),(\mathbf{x}_i,\mathbf{y}_i))\}.
\end{equation}
Here $\mathbf{x}^\star$ can be regarded as an adversarial perturbation of $\mathbf{x}_i$ under current model $\boldsymbol{\phi}_i$ and
efficiently approximated by gradient-ascent methods, revealing that problem \eqref{robustrelax} can be efficiently solved by gradient-based methods. 

\subsection{Robust Meta-Training across Edge Nodes}

To solve problem \eqref{robustrelax}, similar to \cite{volpi2018generalizing}, we use an adversarial data generation process, i.e., approximately solving \eqref{optimalx} with gradient ascent, to the federated meta-learning algorithm. More specifically, for every $N_0T_0$ iterations, each edge node $i$ constructs adversarial data samples using $T_a$ steps gradient ascent and adds them to its own adversarial dataset $D_i^{adv}$. Note that this sample construction procedure can only be repeated up to $R$ times considering the local computational resources constraints.
For $t\neq nT_0$ (no global aggregation), each node $i$ first updates $\boldsymbol{\theta}_i^t$ using the training dataset: 
\begin{equation}\label{updatephi}
    \boldsymbol{\phi}_i^{t}=\boldsymbol{\theta}_i^t-\alpha \nabla_{\boldsymbol{\theta}}L(\boldsymbol{\theta}_i^t, D_i^{train}),
\end{equation}
then locally updates $\boldsymbol{\theta}_i^t$ again using both the testing dataset and the constructed adversarial dataset:
\begin{equation}\label{robust}
    \boldsymbol{\theta}_i^{t+1}=\boldsymbol{\theta}_i^t-\beta\nabla_{\boldsymbol{\theta}}\{L(\boldsymbol{\phi}_i^t, D_i^{test})+L(\boldsymbol{\phi}_i^t, D_i^{adv})\}.
\end{equation}
When $t=nT_0$, each node transmits the updated $\boldsymbol{\theta}_i^{t+1}$ for the global aggregation \eqref{aggregation}. The details are summarized in Algorithm 2.

\begin{algorithm}[!t]
    \footnotesize
	\caption{Robust FedML}\label{alg3}
	\begin{algorithmic}[1]
		  \Statex \textbf{Inputs:} $K$, $T$, $T_0$, $T_a$, $N_0$, $R$, $\alpha$, $\beta$, $\nu$, $\omega_i$ for $i\in\mathcal{S}$\;
		  \Statex \textbf{Outputs:} Final model parameter $\boldsymbol{\theta}$\;
		  \State Platform randomly initializes $\boldsymbol{\theta}^0$ and sends it to all nodes in $\mathcal{S}$; For each node $i$, $D_i^{adv}\leftarrow \emptyset$ and $r\leftarrow 0$;
		\For{$t= 1, 2, ..., T$}
		    \For{each node $i\in\mathcal{S}$}
		         \State $D_i^{comb}\leftarrow D_i^{test}\cup D_i^{adv}$;
		         \State Compute the updated parameter with one-step gradient descent using $D_i^{train}$: $\boldsymbol{\phi}_i^{t}=\boldsymbol{\theta}_i^t-\alpha \nabla_{\boldsymbol{\theta}}L(\boldsymbol{\theta}_i^t, D_i^{train})$;
		        \State Obtain $\boldsymbol{\theta}_i^{t+1}$ based on \eqref{robust};
		        \textbf{//local update}
		        \If{$t$ mod $T_0=0$}
		          \State Send $\boldsymbol{\theta}_i^{t+1}$ to the platform;
		          \State Receive $\boldsymbol{\theta}^{t+1}$ from the platform where $\boldsymbol{\theta}^{t+1}$ is obtained based on \eqref{aggregation};
		          \State Set $\boldsymbol{\theta}_i^{t}\leftarrow \boldsymbol{\theta}^{t+1}$; \textbf{//global aggregation}
		         \Else
		         \State Set $\boldsymbol{\theta}_i^{t}\leftarrow \boldsymbol{\theta}_i^{t+1}$;
		      \EndIf
		      \If{$t$ mod $N_0T_0=0$ and $r<R$}
		          \textbf{//adversarial data generation}
		          \State Uniformly sample   $(\mathbf{x}_i^j,\mathbf{y}_i^j)_{j=1,...,|D_i^{test}|}$ from $D_i^{comb}$\;
		          \For{$j=1,...,|D_i^{test}|$}
		                \State $\mathbf{x}_i^{jr}\leftarrow \mathbf{x}_i^j$;
		                \For{$t=1,...,T_a$}
		                    \State $\mathbf{x}_i^{jr}\leftarrow \mathbf{x}_i^{jr}+\nu\nabla_x\{l(\boldsymbol{\phi}_i^t,(\mathbf{x}_i^{jr},\mathbf{y}_i^j))-\lambda c((\mathbf{x}_i^{jr},\mathbf{y}_i^j),(\mathbf{x}_i^{j},\mathbf{y}_i^j))\}$;
		                \EndFor
		                \State Append $(\mathbf{x}_i^{jr},\mathbf{y}_i^j)$ to $D_i^{adv}$;
		          \EndFor
		          \State $r\leftarrow r+1$;
		      \EndIf
		    \EndFor
		 \EndFor
		 \State \textbf{return} $\boldsymbol{\theta}$.
	\end{algorithmic}
\end{algorithm}

\subsection{Convergence Analysis}
Similar with Section 4.1, for clarity we rewrite problem \eqref{robustrelax} as the following:
\begin{mini}
{\boldsymbol{\theta}}{\Tilde{G}(\boldsymbol{\theta}),}
{\label{tildeG}}{}
\end{mini}
where $\Tilde{G}(\boldsymbol{\theta})=\sum_{i\in\mathcal{S}}\omega_i\Tilde{G}_i(\boldsymbol{\theta})$ and $\Tilde{G}_i(\boldsymbol{\theta})=L(\boldsymbol{\phi}_i(\boldsymbol{\theta}),D_i^{test})+\mathbb{E}_{P_i}[l_{\lambda}(\boldsymbol{\phi}_i(\boldsymbol{\theta}),(\mathbf{x}_i,\mathbf{y}_i))]$. 
\begin{assumption}
The function $c$ is continuous. And for every $(\mathbf{x}_0,\mathbf{y}_0)\in \mathcal{X}\times\mathcal{Y}$, $c((\mathbf{x},\mathbf{y}_0),(\mathbf{x}_0,\mathbf{y}_0))$ is 1-strongly convex with respect to $\mathbf{x}$.
\end{assumption}
\begin{assumption}
The loss function $l:\mathbb{R}^d\times (\mathcal{X}\times\mathcal{Y})\rightarrow\mathbb{R}$ is $\mu$-strongly convex with respect to $\boldsymbol{\theta}$.
\end{assumption}
\begin{assumption}
The loss function $l:\mathbb{R}^d\times (\mathcal{X}\times\mathcal{Y})\rightarrow\mathbb{R}$ is smooth with respect to both $\boldsymbol{\theta}$ and $\mathbf{x}$, i.e.,
\begin{align*}
    &\|\nabla_{\boldsymbol{\theta}}l(\boldsymbol{\theta},(\mathbf{x},\mathbf{y}))-\nabla_{\boldsymbol{\theta}}l(\boldsymbol{\theta}',(\mathbf{x},\mathbf{y}))\|\leq H\|\boldsymbol{\theta}-\boldsymbol{\theta}'\|,\\
    &\|\nabla_{\boldsymbol{\theta}}l(\boldsymbol{\theta},(\mathbf{x},\mathbf{y}))-\nabla_{\boldsymbol{\theta}}l(\boldsymbol{\theta},(\mathbf{x}',\mathbf{y}))\|\leq H_{\boldsymbol{\theta}\mathbf{x}}\|\mathbf{x}-\mathbf{x}'\|,\\
    &\|\nabla_{\mathbf{x}}l(\boldsymbol{\theta},(\mathbf{x},\mathbf{y}))-\nabla_{\mathbf{x}}l(\boldsymbol{\theta},(\mathbf{x}',\mathbf{y}))\|\leq H_{\mathbf{x}\mathbf{x}}\|\mathbf{x}-\mathbf{x}'\|,\\
    &\|\nabla_{\mathbf{x}}l(\boldsymbol{\theta},(\mathbf{x},\mathbf{y}))-\nabla_{\mathbf{x}}l(\boldsymbol{\theta}',(\mathbf{x},\mathbf{y}))\|\leq H_{\mathbf{x}\boldsymbol{\theta}}\|\boldsymbol{\theta}-\boldsymbol{\theta}'\|,
    \vspace{-0.2cm}
\end{align*}
and there exists a constant $B$ such that $\|\nabla_{\boldsymbol{\theta}}l(\boldsymbol{\theta},(\mathbf{x},\mathbf{y}))\|\leq B$ for all $\boldsymbol{\theta}\in\mathbb{R}^d$ and $(\mathbf{x},\mathbf{y})\in\mathcal{X}\times\mathcal{Y}$.
\end{assumption}
Note that Assumptions 6-7 can be stronger replacements of Assumptions 1-2, respectively. We characterize the robust federated meta-learning objective $\Tilde{G}(\boldsymbol{\theta})$ below:  
\begin{theorem}
Suppose Assumptions 3 and 5-7 hold. When $\alpha\leq \min\{\frac{\mu}{2\mu H+\rho B},\frac{1}{\mu}\}$ and $\lambda\geq H_{\mathbf{xx}}+\frac{H_{\boldsymbol{\theta}\mathbf{x}}H_{\mathbf{x}\boldsymbol{\theta}}}{\mu}$,
problem \eqref{tildeG} has a unique minimizer.
\end{theorem}
Theorem 4 implies that when the learning rate $\alpha$ is sufficiently small and the Lagrangian penalty parameter $\lambda$ is large enough, the relaxed robust meta-learning objective function $\Tilde{G}(\boldsymbol{\theta})$ is strongly convex and hence has a unique solution. Further, as outlined in Algorithm 2, through pre-training each edge node to learn to protect against adversarial perturbations on the testing dataset while securing the model accuracy on clean data with Algorithm 2, the learned model via meta-training automatically gains the ability to prevent future adversarial attacks without significantly sacrificing the learning accuracy with quick adaptation at the target edge node.

\section{Experiments}
In this section, we first introduce the experimental setup, and then
evaluate the performance of FedML and Robust FedML. In particular, we investigate the impact of node similarity and number of local update step $T_0$ on the convergence of federated meta-learning, and compare the fast adaptation performance of federated meta-learning (FedML) with the model learnt from federated learning (Fedavg) \cite{mcmahan2016communication}. 

\subsection{Experimental Setting}
\textbf{Synthetic data.} To evaluate the impact of node similarity on the performance of convergence and fast adaptation, we follow a similar setup in \cite{sahu2018convergence} to generate synthetic data. Specifically, the synthetic sample $(\mathbf{x}_i^j, \mathbf{y}_i^j)$ for each node $i$ is generated from the model $\mathbf{y}=argmax(softmax(\mathbf{Wx+b}))$ where $\mathbf{x}\in\mathbb{R}^{60}$, $\mathbf{W}\in\mathbb{R}^{10\times 60}$ and $\mathbf{b}\in\mathbb{R}^{10}$. Moreover, $\mathbf{W}_i\sim \mathcal{N}(\mathbf{u}_i,\mathbf{1})$, $\mathbf{b}_i\sim\mathcal{N}(\mathbf{u}_i,\mathbf{1})$, $\mathbf{u}_i\sim\mathcal{N}(0,\Tilde{\alpha})$; $\mathbf{x}_i^j\sim\mathcal{N}(\mathbf{v}_i, \Sigma)$ where the covariance matrix $\Sigma$ is diagonal with $\Sigma_{k,k}=k^{-1.2}$ and $\mathbf{v}_i\sim\mathcal{N}(B_i,1)$, $B_i\sim N(0,\Tilde{\beta})$. Intuitively, $\Tilde{\alpha}$ and $\Tilde{\beta}$ control the local model similarity across all nodes, which can be changed to generate heterogeneous local datasets named Synthetic($\Tilde{\alpha}$, $\Tilde{\beta}$). For all synthetic datasets, we consider 50 nodes in total and the number of samples on each node follows a power law. The objective is to learn the model parameters $\mathbf{W}$ and $\mathbf{b}$ with the cross-entropy loss function.

\textbf{Real data.} We also explore two real datasets, MNIST \cite{lecun1998gradient}, and Sentiment140 (Sent140) \cite{go2009twitter} used for text sentiment analysis on tweets. For MNIST, we sample part of data and distribute the data among 100 nodes such that every node has samples of only two digits and the number of samples per device follows a power law. We study a convex classification problem with MNIST using multinomial logistic regression. Next, we consider a more complicated classification problem on Sent140 by taking each twitter account as a node, where the model takes a sequence of 25 characters as input, embeds each of the character into a 300 dimensional space by looking up the pretrained 300D GloVe embedding \cite{pennington2014glove}, and outputs one character per training sample through a network with 3 hidden layers with sizes 256, 128, 64, each including batch normalization and ReLU nonlinearities, followed by a linear layer and softmax. The loss function is the cross-entropy error between the predicted and true class for all models.
 \begin{table}[t]\label{tab1}
 \caption{Statistics of Datasets}
 \small
 \vspace*{-2ex}
 \begin{center}
\begin{tabular}{ llll }
  \multirow{2}{*}{\textbf{Dataset}}& \multirow{2}{*}{\textbf{Nodes}} & \textbf{Sample per Node} \\
  \cline{3-4}
  &  & mean & stdev\\
 \hline
 Synthetic & 50 & 17 & 5 \\
 MNIST & 100 & 34 & 5\\
Sent140 & 706  & 42 & 35\\
\end{tabular}
\end{center}
\vspace*{-3.5ex}
\end{table}

\textbf{Implementation.} For each node, we divide the local dataset as a training set and a testing set. We select 80\% nodes as the source nodes and evaluate the fast adaptation performance on the rest. When training with FedML, we vary the size of the training set, i.e., $K$, for the one-step gradient update, whereas the entire dataset is used for training in Fedavg. During testing, the trained model is first updated with the training set of testing nodes, and then evaluated on their testing sets. For FedML, we set both the learning rate $\alpha$ and meta learning rate $\beta$ as 0.01 for synthetic data and MNIST, while $\alpha=0.01$ and $\beta=0.3$ for Sent140. Fedavg has the same learning rate with $\beta$. 

\subsection{Evaluation of Federated Meta-Learning}
\textbf{Convergence behaviour.}  We evaluate the convergence error for (a) three different synthetic datasets with $T_0=10$ (b) the same dataset but different $T_0$ with $T=500$. As illustrated in Figure 2, the experimental results corroborate Theorem 2 that the convergence error of FedML decreases with the node similarity but increases with $T_0$ given a fixed algorithm duration $T$.  Moreover, the result in Sent140 (Figure 3(a)) shows that FedML also achieves good convergence performance in practical non-convex settings.
\begin{figure}
    \centering
    \begin{subfigure}[b]{0.5\textwidth}
    \includegraphics[width=0.95\textwidth]{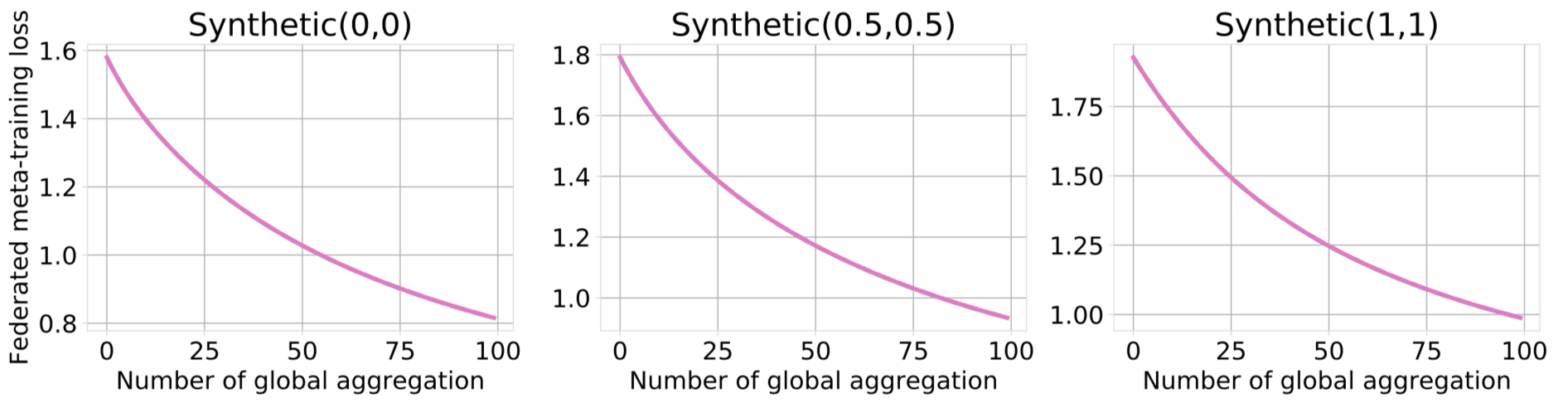}
    \caption{Impact of node similarity}
    \end{subfigure}%
    
    \begin{subfigure}[b]{0.5\textwidth}
    \includegraphics[width=0.95\textwidth]{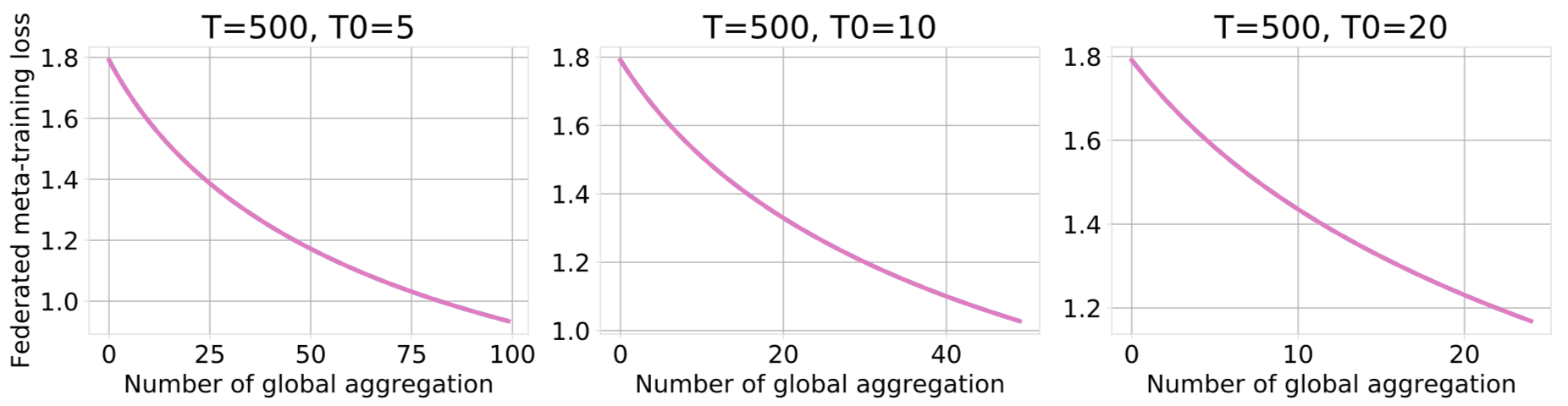}
    \caption{Impact of $T_0$ with Synthetic(0.5,0.5)}
    \end{subfigure}%
    \caption{Impact of Node Similarity and $T_0$ on the Convergence of FedML}
    \label{fig:impact_conv}
    \vspace{-0.6cm}
\end{figure}

\textbf{Fast adaptation performance.} As shown in Figure 3(b), FedML achieves the best adaptation performance on Synthetic(0,0) where the nodes are the most similar. We also compare the fast adaptation performance between FedML and Fedavg on three different datasets, Synthetic(0.5,0.5), MNIST and Sent140. As shown in Figure 3(c)-3(e), the model learnt from FedML can achieve significantly better adaptation performance at the target nodes compared with that in Fedavg, and this performance gap increases when the target node has small local datasets. It can be seen that the model learnt in Fedavg turns to have overfitting issues when fine-tuned with a few data samples, whereas the meta-model in FedML improves with additional gradient steps without overfitting.
\begin{figure*}
    \centering
    \begin{subfigure}[b]{0.195\textwidth}
        \includegraphics[width=1.0\linewidth]{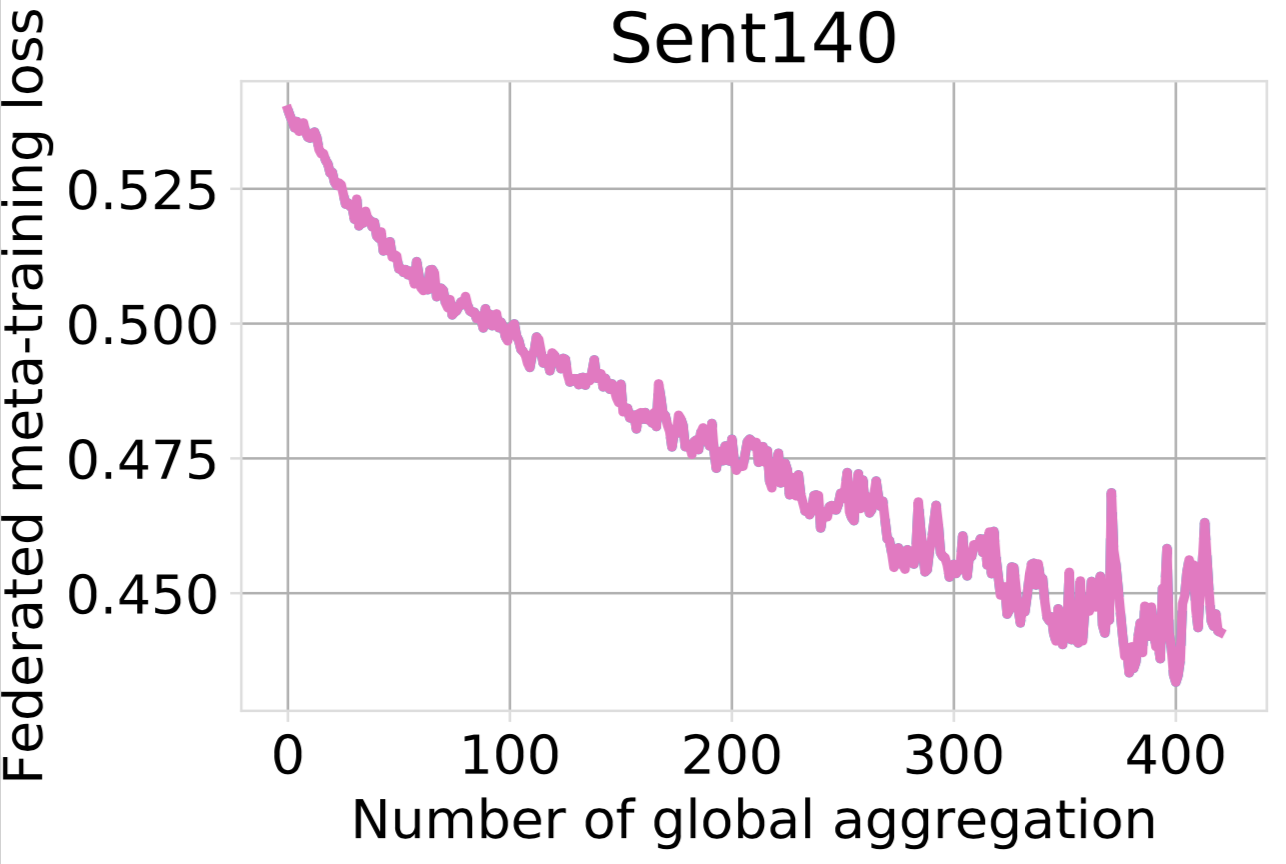}
        \caption{Convergence of FedML on Sent140}
    \end{subfigure}
    \begin{subfigure}[b]{0.195\textwidth}
        \includegraphics[width=1.0\linewidth]{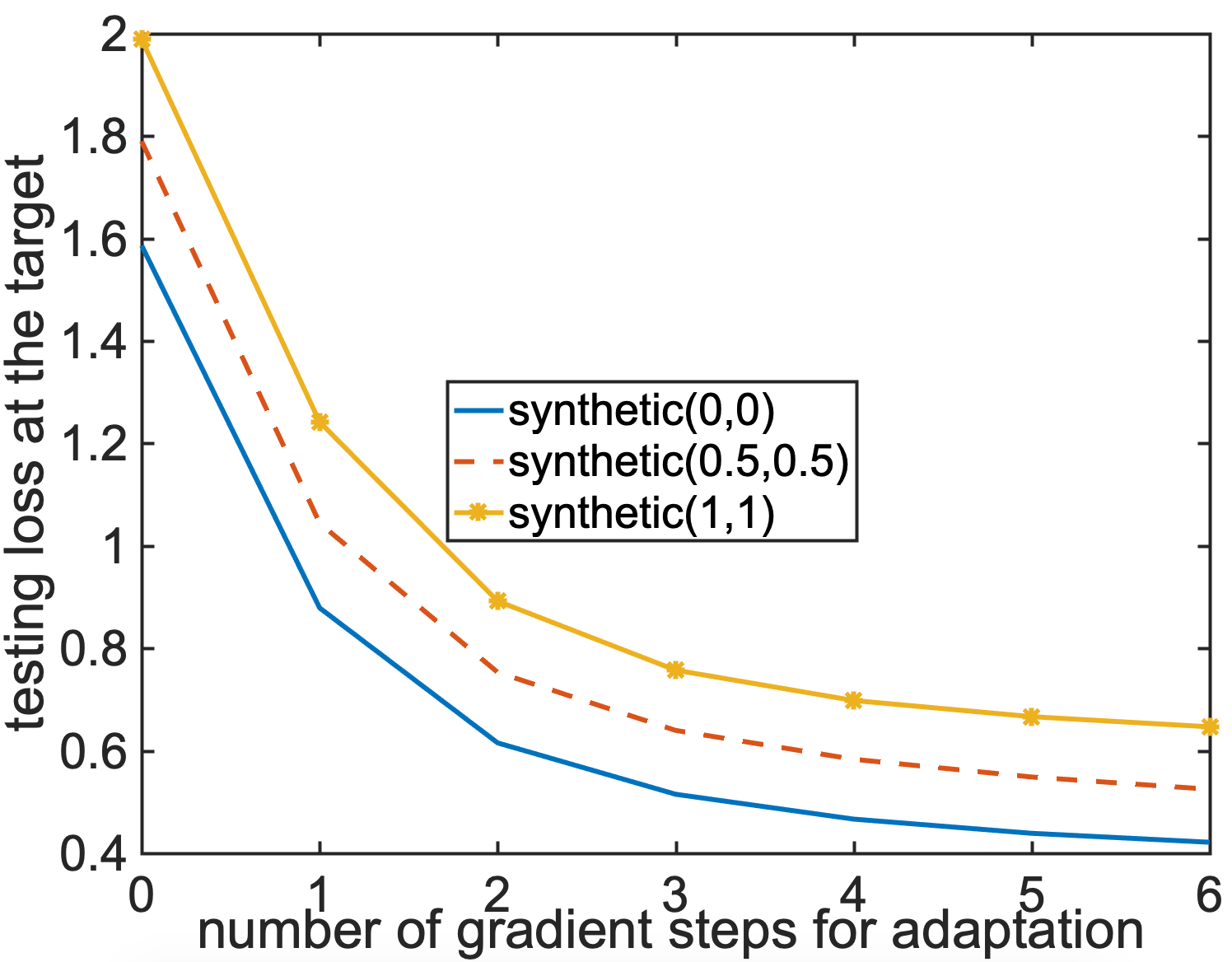}
        \caption{Impact of target-source similarity on test performance}
    \end{subfigure}
    \begin{subfigure}[b]{0.195\textwidth}
        \includegraphics[width=1.0\linewidth]{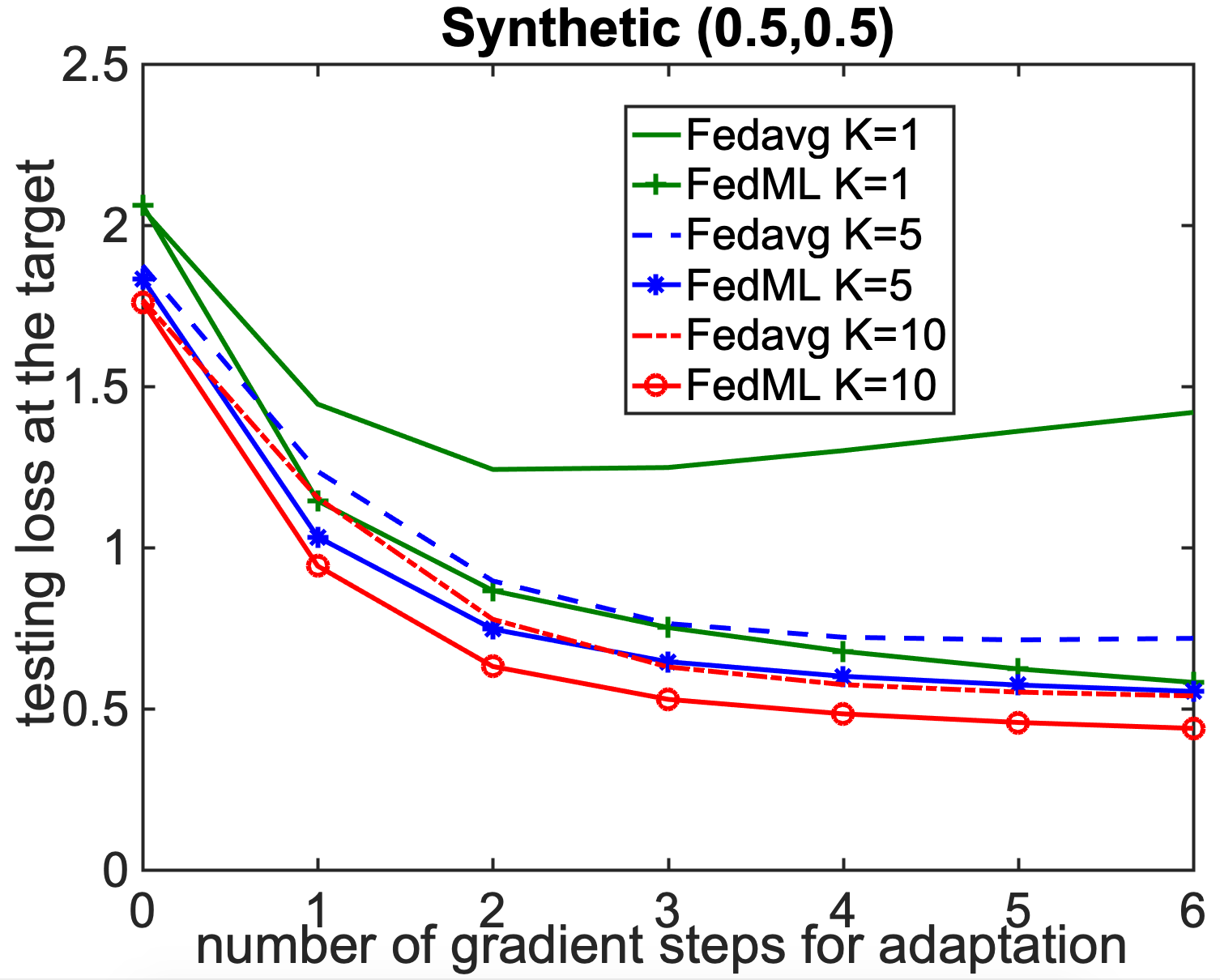}
        \caption{Adaptation performance on Synthetic(0.5,0.5)}
    \end{subfigure}
    \begin{subfigure}[b]{0.195\textwidth}
        \includegraphics[width=1.0\linewidth]{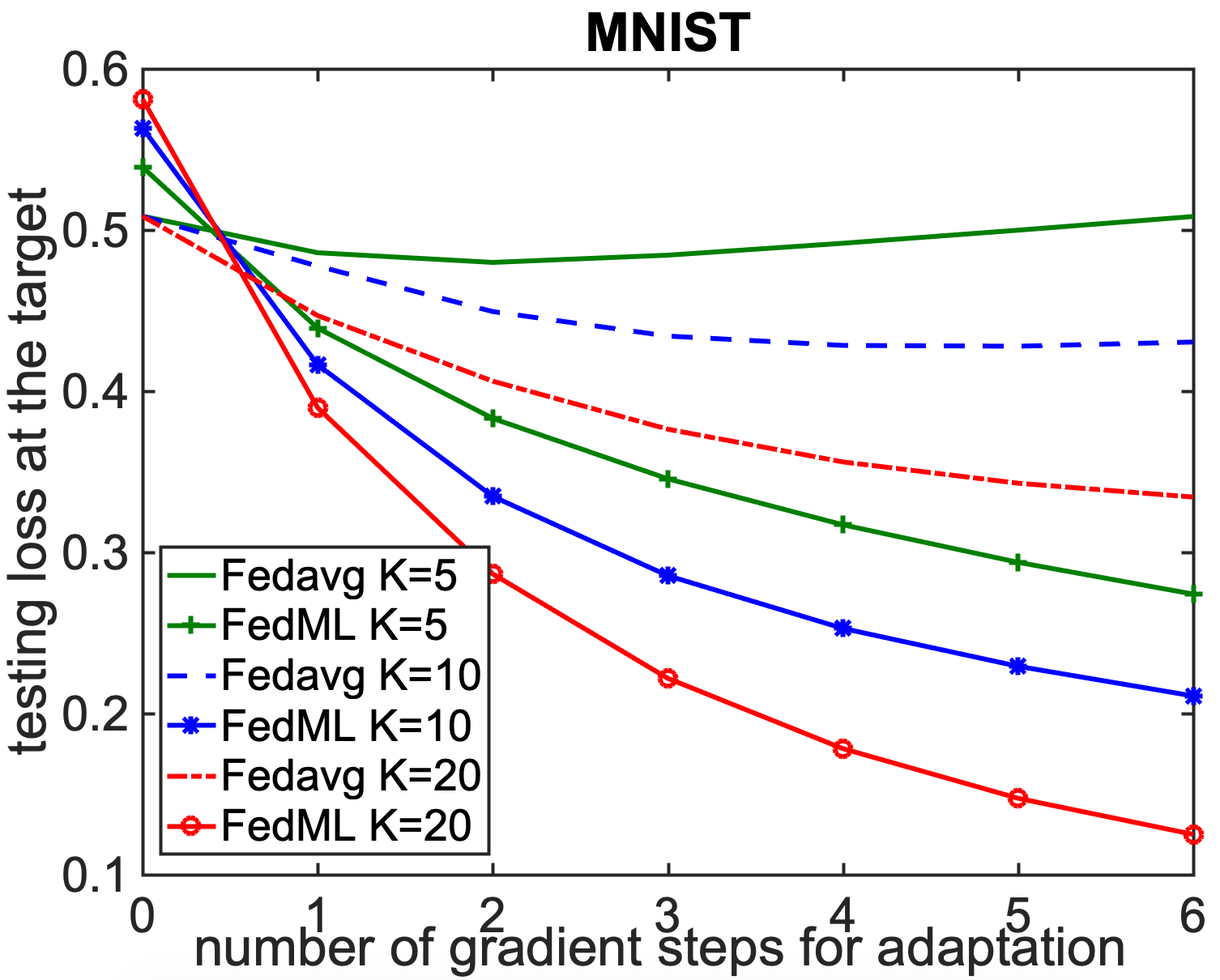}
        \caption{Adaptation performance on MNIST}
    \end{subfigure}
    \begin{subfigure}[b]{0.195\textwidth}
        \includegraphics[width=1.0\linewidth]{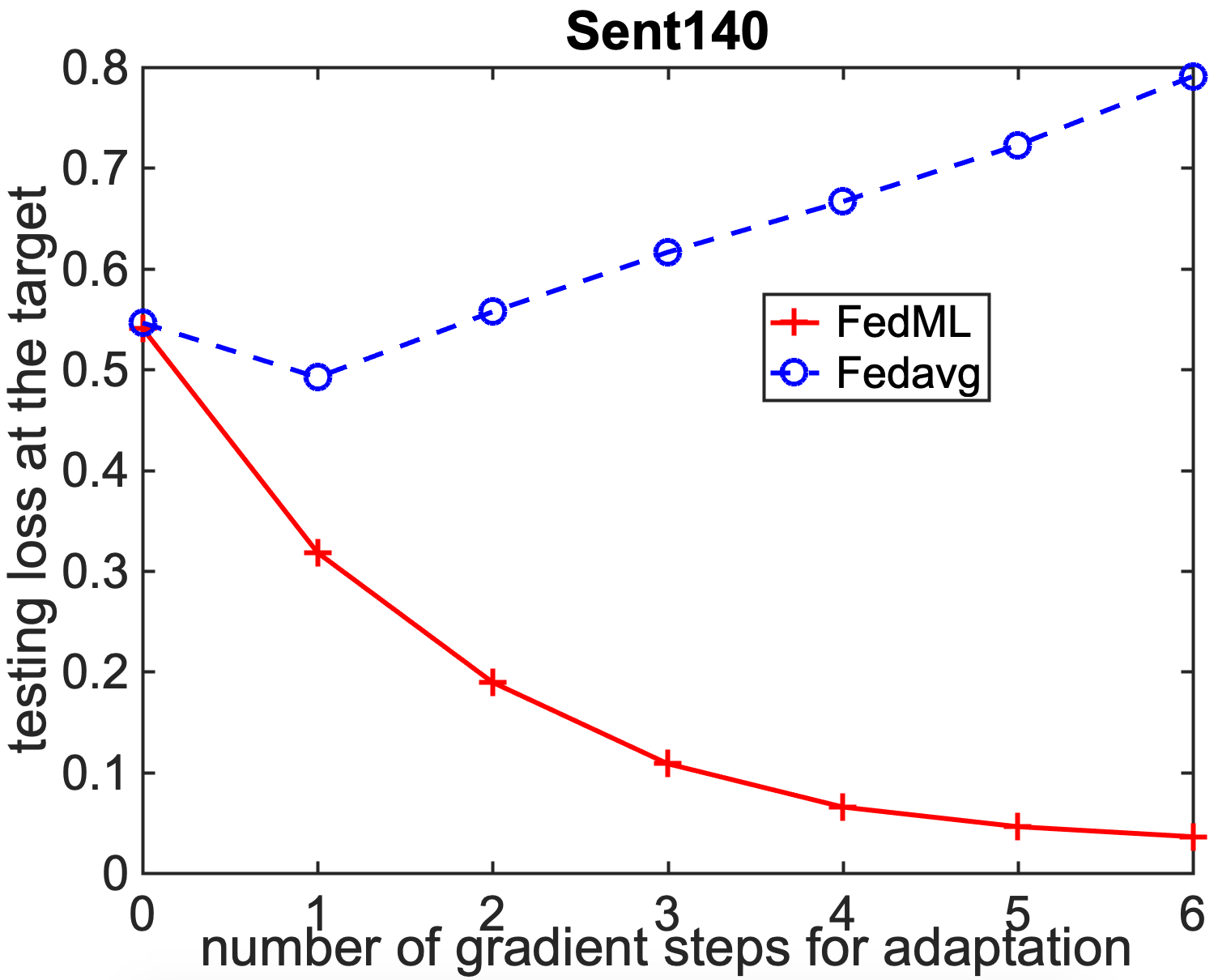}
        \caption{Adaptation performance on Sent140}
    \end{subfigure}
    \caption{Convergence and Fast Adaptation Performance of FedML on Different Datasets with $T_0=5$}
    \label{fig:impact_conv}
    \vspace{-0.5cm}
\end{figure*}
\subsection{Evaluation of Robust Federated Meta-Learning}
\begin{figure*}
    \centering
    \begin{subfigure}[b]{0.195\textwidth}
        \includegraphics[width=1.0\linewidth]{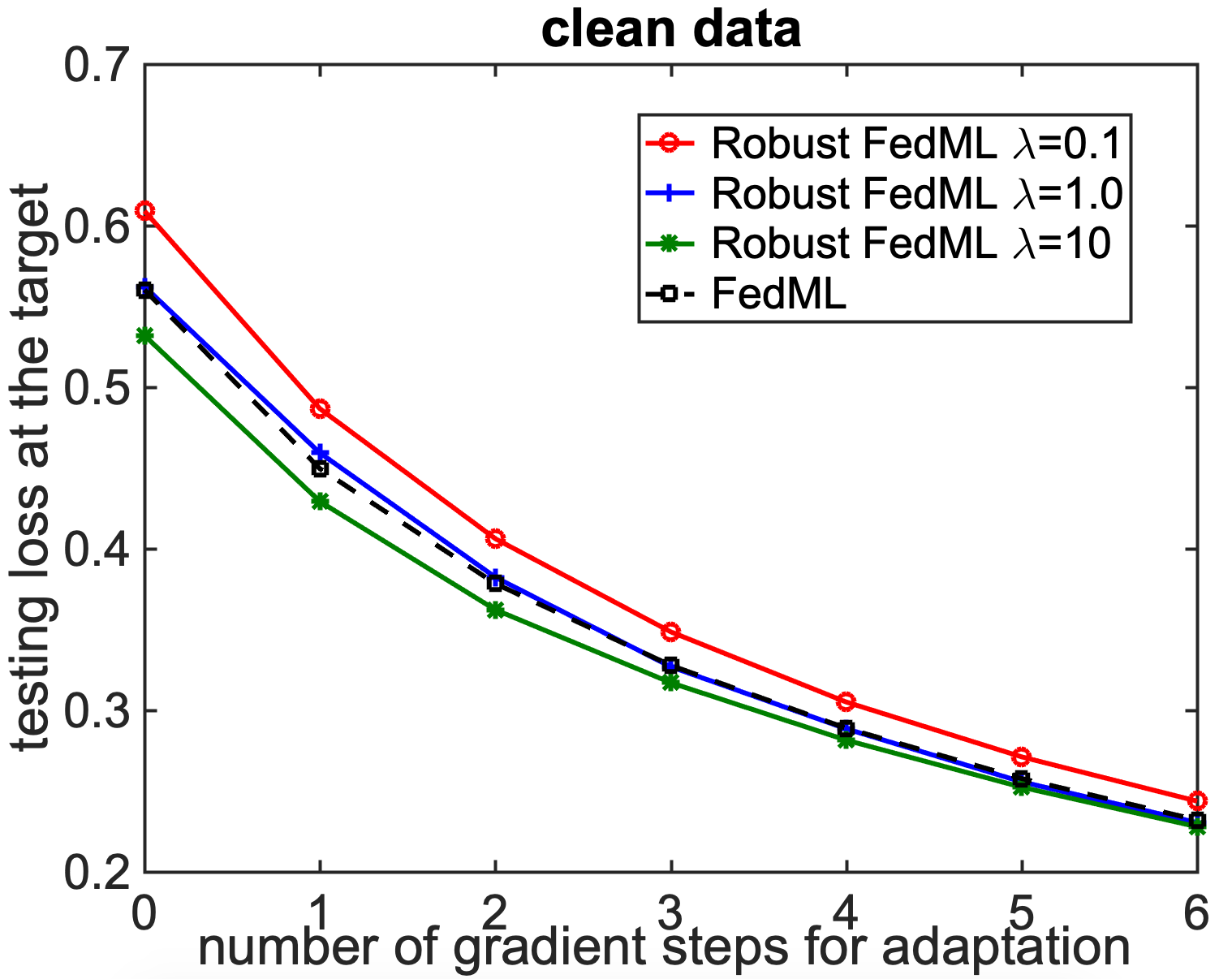}
        \caption{Loss on clean data}
    \end{subfigure}
        \begin{subfigure}[b]{0.195\textwidth}
        \includegraphics[width=1.0\linewidth]{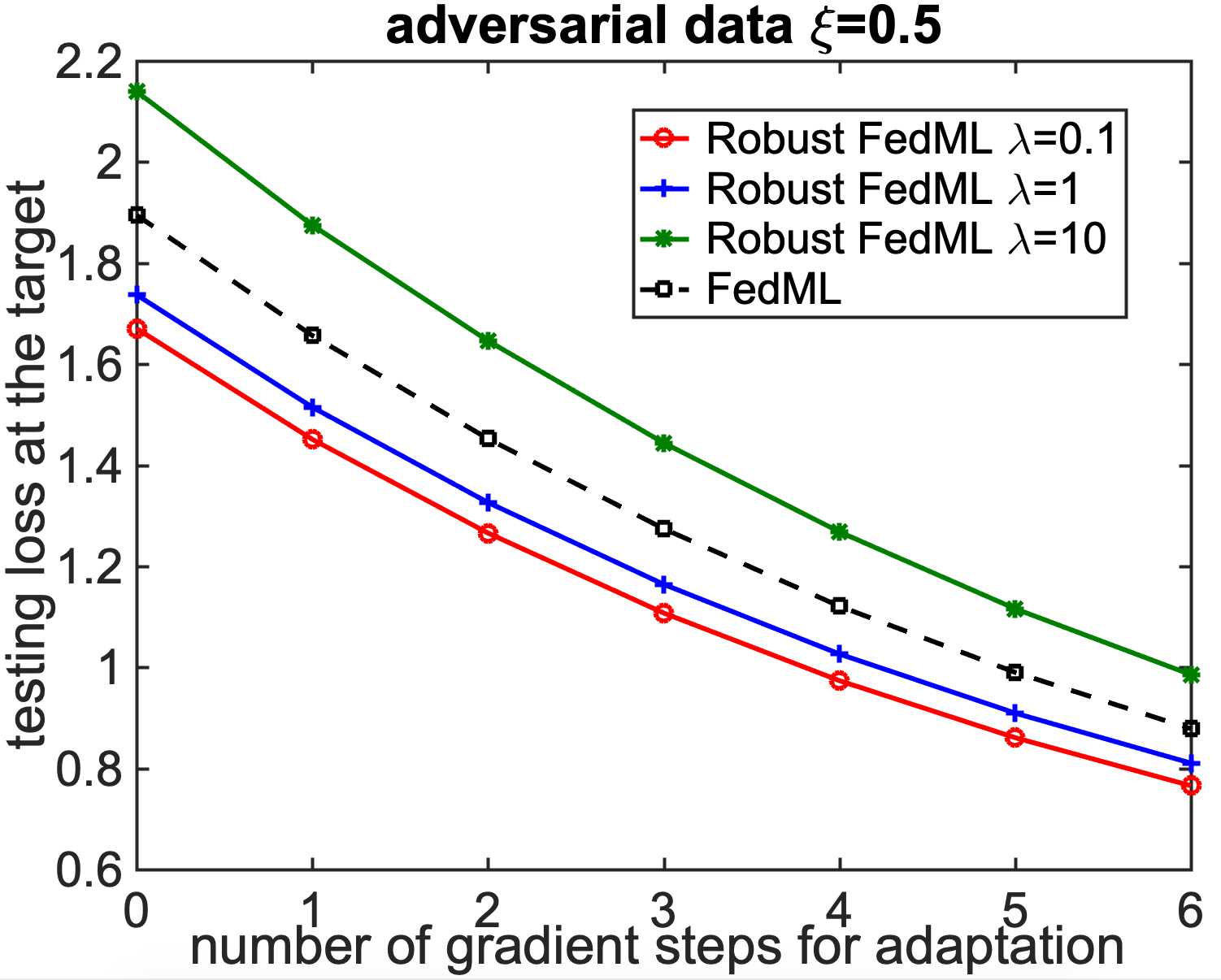}
        \caption{Loss on adversarial data}
    \end{subfigure}
        \begin{subfigure}[b]{0.195\textwidth}
        \includegraphics[width=1.0\linewidth]{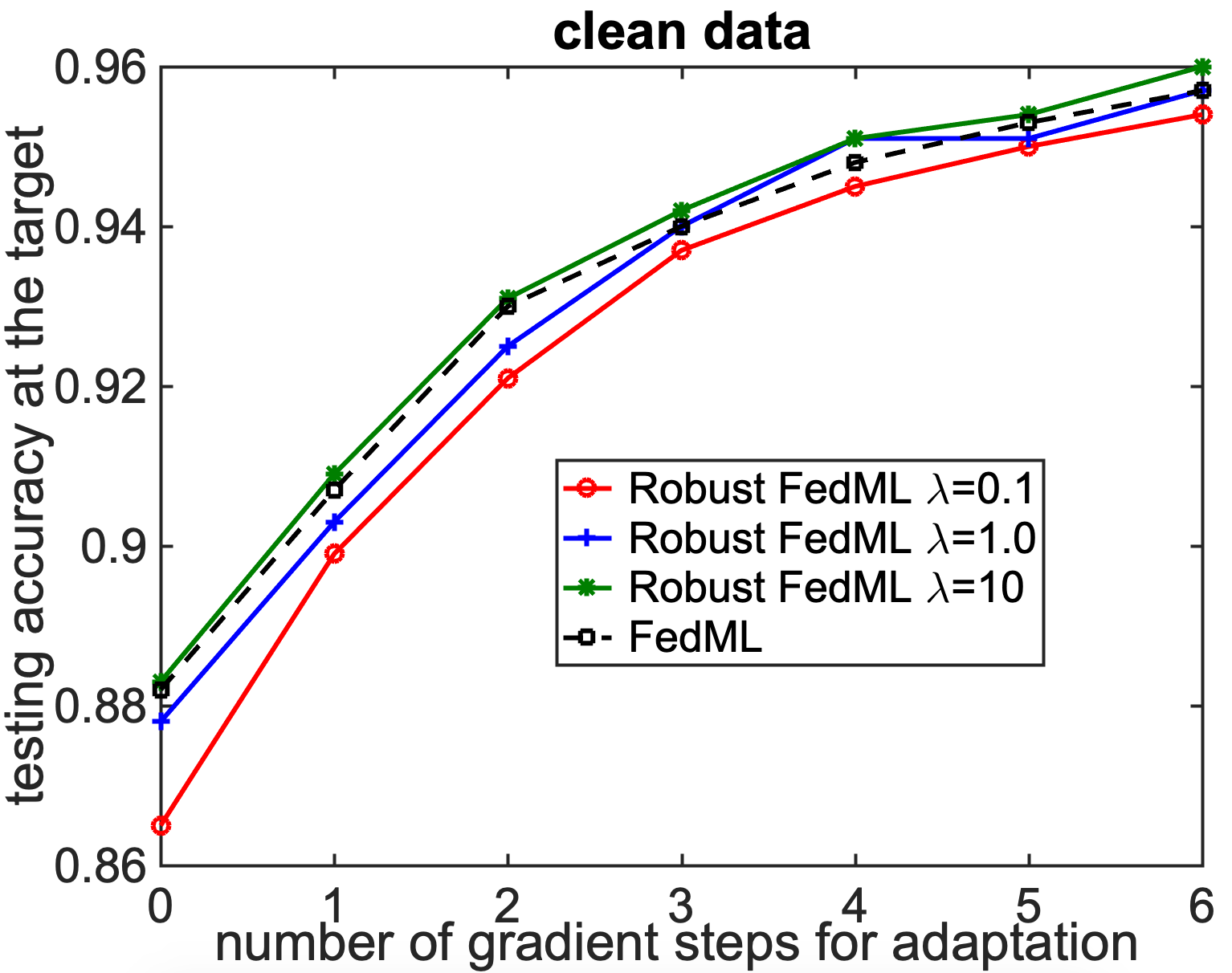}
        \caption{Accu. on clean data}
    \end{subfigure}
    \begin{subfigure}[b]{0.195\textwidth}
        \includegraphics[width=1.0\linewidth]{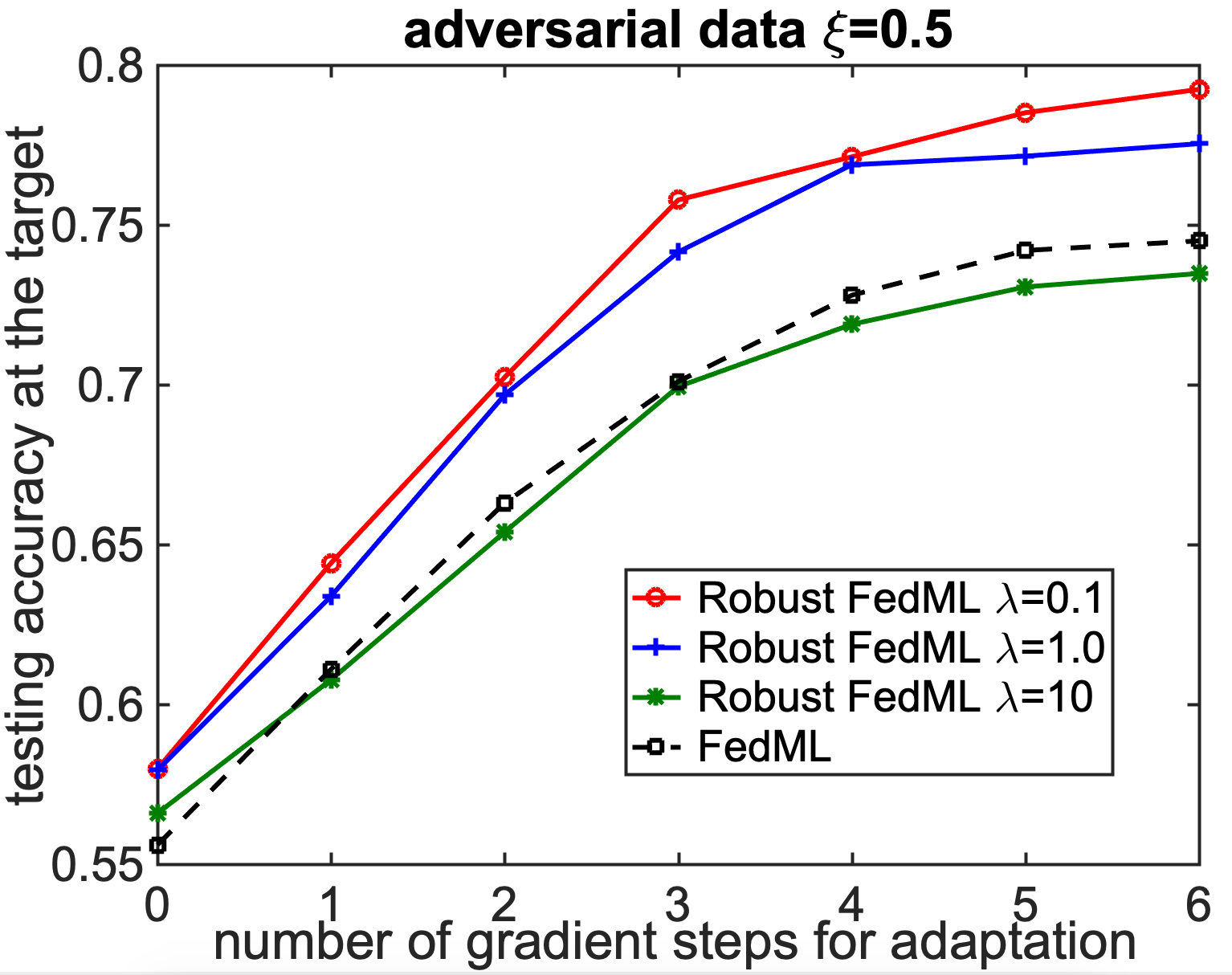}
        \caption{Accu. on adversarial data}
    \end{subfigure}
    \begin{subfigure}[b]{0.195\textwidth}
        \includegraphics[width=1.0\linewidth]{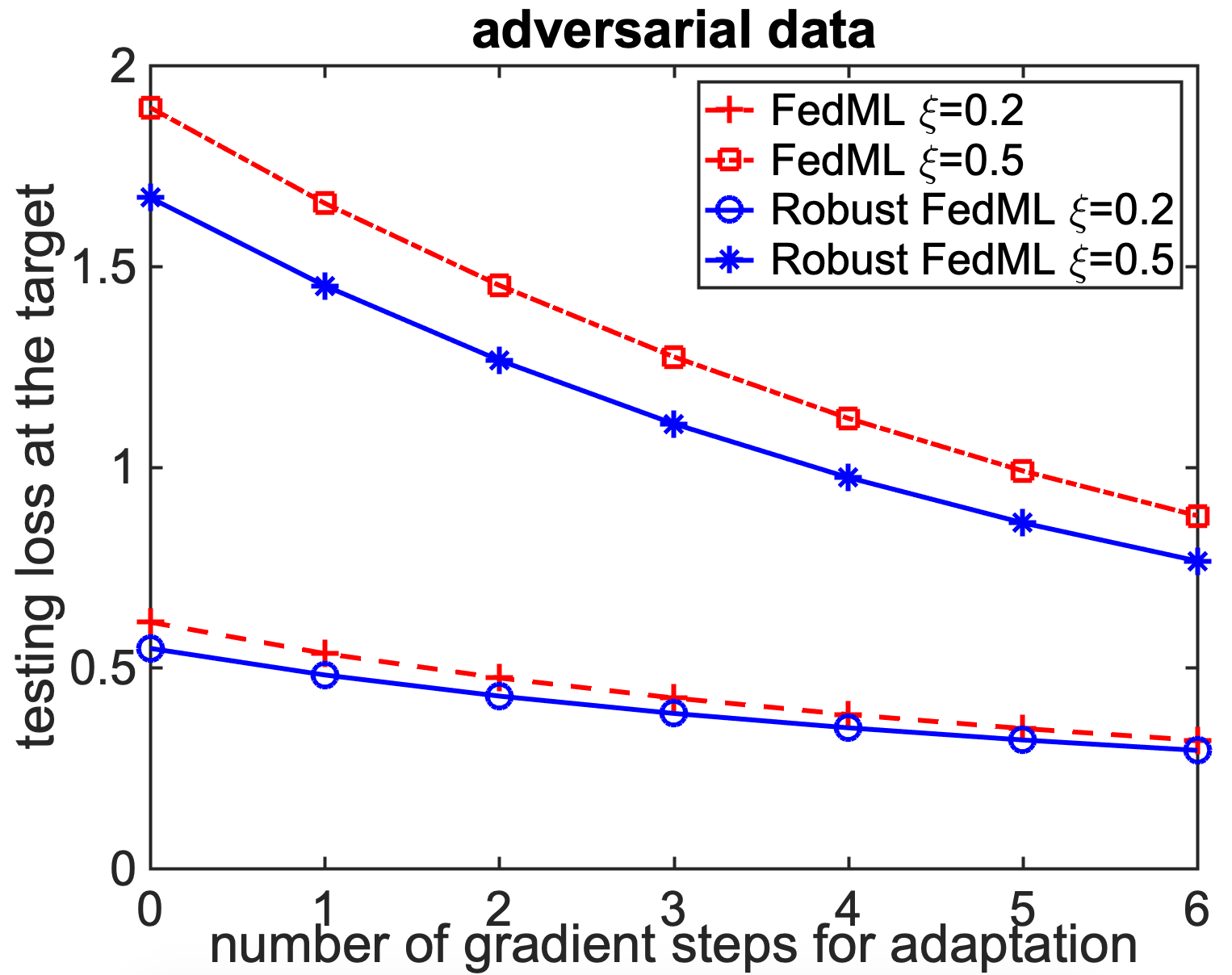}
        \caption{Impact of $\xi$}
    \end{subfigure}
    \caption{Adaptation Performance of Robust FedML on MNIST with $T_0=5$}
    \label{fig:impact_conv}
    \vspace{-0.7cm}
\end{figure*}
We compare the adaptation performance of FedML and Robust FedML on MNIST with $T_0=5$ during training. For adversarial perturbations only to feature vectors in supervised learning, we consider the transportation cost function as: $c((\mathbf{x},\mathbf{y}),(\mathbf{x}',\mathbf{y}'))=\|\mathbf{x}-\mathbf{x}'\|_2^2+\infty\cdot\mathbf{1}(\mathbf{y}-\mathbf{y}')$. The learning rate $\nu=1$, $R=2$, $N_0=7$ and $T_a=10$ for adversarial data generation. For testing at the target, we first update the meta-model with clean training data, and then evaluate the adaptation performance on clean test data and adversarial data, where the adversarial data is generated by using the Fast Gradient Sign Method \cite{goodfellow2014explaining} with parameter $\xi$, respectively. Since the size of distributional uncertainty set controls the trade-off between robustness and accuracy, we compare the performance of Robust FedML with $\lambda=0.1, 1, 10$, where the smaller $\lambda$ is, the more robustness Robust FedML provides.

\textbf{Robustness-Accuracy tradeoff.} As shown in Figure 4(a)-4(d),
when $\lambda$ decreases, Robust FedML has slightly worse performance on clean data and the performance on adversarial data is much better. Compared with the case where $\lambda=10$, Robust FedML with $\lambda=0.1$ significantly improves the robustness against adversarial data without sacrificing too much on the accuracy with clean data. Moreover, Robust FedML with smaller $\lambda$ is more robust than FedML. Note that the uncertainty set is too small to positively affect the robustness when $\lambda=10$.

\textbf{Impact of $\xi$}. Clearly, both FedML and Robust FedML achieve better performance when facing smaller perturbation (smaller $\xi$) of testing data. Figure 4(e) further indicates that the improvement of Robust FedML over FedML is higher with more perturbed data.

\section{CONCLUSION}
In this paper, we propose a platform-aided collaborative learning framework, where a model is first trained across a set of source edge nodes by a federated meta-learning approach, and then it is rapidly adapted to  achieve real-time edge intelligence at the target edge node, using a few samples only. We investigate the convergence of FedML under mild conditions on node similarity , and study the adaptation performance to achieve edge intelligence at the target node.  To combat against the vulnerability of meta-learning algorithms, we further propose a robust FedML algorithm based on DRO with convergence guarantee. Experimental results on various datasets corroborate the effectiveness of the proposed collaborative learning framework.


\bibliography{bibliography}
\bibliographystyle{IEEEtran}

\appendices

\section{Proof of Lemma 1}
We  first show that $G_i(\boldsymbol{\theta})$ is $\mu'$-strongly convex and $L'$-smooth. Specifically, observe that
\begin{small}
     \begin{align}\label{G_i}
         &\|\nabla G_i(\boldsymbol{\theta})-\nabla G_i(\boldsymbol{\theta}')\|\nonumber\\
        =&\|\nabla L_i(\boldsymbol{\phi}_i)-\alpha \nabla^2 L_i(\boldsymbol{\theta})\nabla L_i(\boldsymbol{\phi}_i)-\nabla L_i(\boldsymbol{\phi}'_i)\nonumber\\
         &+\alpha \nabla^2 L_i(\boldsymbol{\theta}')\nabla L_i(\boldsymbol{\phi}'_i)\| \nonumber\\
         =&\|\nabla L_i(\boldsymbol{\phi}_i)-\alpha \nabla^2 L_i(\boldsymbol{\theta})\nabla L_i(\boldsymbol{\phi}_i)+\alpha \nabla^2 L_i(\boldsymbol{\theta}')\nabla L_i(\boldsymbol{\phi}'_i)\nonumber\\
         &-\nabla L_i(\boldsymbol{\phi}'_i)+\alpha\nabla^2 L_i(\boldsymbol{\theta})\nabla L_i(\boldsymbol{\phi}'_i)-\alpha\nabla^2 L_i(\boldsymbol{\theta})\nabla L_i(\boldsymbol{\phi}'_i)\|\nonumber\\
         =&\|[I-\alpha \nabla^2 L_i(\boldsymbol{\theta})][\nabla L_i(\boldsymbol{\phi}_i)-\nabla L_i(\boldsymbol{\phi}'_i)]\nonumber\\
         &-\alpha\nabla L_i(\boldsymbol{\phi}'_i)[\nabla^2 L_i(\boldsymbol{\theta})-\nabla^2 L_i(\boldsymbol{\theta}')]\| ,
     \end{align}
\end{small}
where $\boldsymbol{\phi}_i=\boldsymbol{\theta}-\alpha\nabla L_i(\boldsymbol{\theta})$ and $\boldsymbol{\phi}'_i=\boldsymbol{\theta}'-\alpha\nabla L_i(\boldsymbol{\theta}')$.

To establish the convexity, it suffices to show $\|\nabla G_i(\boldsymbol{\theta})-\nabla G_i(\boldsymbol{\theta}')\|\geq \mu'\|\boldsymbol{\theta}-\boldsymbol{\theta}'\|$. It can be seen from \eqref{G_i} that
\begin{align} \label{convex}
     &\|\nabla G_i(\boldsymbol{\theta})-\nabla G_i(\boldsymbol{\theta}')\|\nonumber\\
     \geq& (1-\alpha H)\|\nabla L_i(\boldsymbol{\phi}_i)-\nabla L_i(\boldsymbol{\phi}'_i)\|\nonumber\\
     &-\alpha\|\nabla L_i(\boldsymbol{\phi}'_i)\|\|\nabla^2 L_i(\boldsymbol{\theta})-\nabla^2 L_i(\boldsymbol{\theta}')\|\nonumber\\
     \geq& \mu(1-\alpha H)\|\boldsymbol{\phi}_i-\boldsymbol{\phi}'_i\|-\alpha\rho B\|\boldsymbol{\theta}-\boldsymbol{\theta}'\|.
\end{align}
Since $\nabla \boldsymbol{\phi}_i=I-\alpha\nabla^2 L_i(\boldsymbol{\theta})$, it follows from Assumption 1 and 2 that  $1-\alpha H\leq \nabla \boldsymbol{\phi}_i\leq 1-\alpha\mu$, which indicates
\begin{equation}\label{phi}
     (1-\alpha H)\|\boldsymbol{\theta}-\boldsymbol{\theta}'\|\leq \|\boldsymbol{\phi}_i-\boldsymbol{\phi}'_i\|\leq (1-\alpha\mu)\|\boldsymbol{\theta}-\boldsymbol{\theta}'\|.
\end{equation}
Combining \eqref{convex} and \eqref{phi}, we have
\begin{equation*}
     \|\nabla G_i(\boldsymbol{\theta})-\nabla G_i(\boldsymbol{\theta}')\|
     \geq \mu'\|\boldsymbol{\theta}-\boldsymbol{\theta}'\|,
\end{equation*}
where $\mu'=\mu(1-\alpha H)^2-\alpha\rho B>0$.

To establish the smoothness, it suffices to show $\|\nabla G_i(\boldsymbol{\theta})-\nabla G_i(\boldsymbol{\theta}')\|\leq H'\|\boldsymbol{\theta}-\boldsymbol{\theta}'\|$. From \eqref{G_i} and \eqref{phi}, we have
     \begin{align*}
         &\|\nabla G_i(\boldsymbol{\theta})-\nabla G_i(\boldsymbol{\theta}')\|\\
         \leq&(1-\alpha\mu)\|\nabla L_i(\boldsymbol{\phi}_i)
         -\nabla L_i(\boldsymbol{\phi}'_i)\|\\
        &+\alpha\|\nabla L_i(\boldsymbol{\phi}'_i)\|\|\nabla^2 L_i(\boldsymbol{\theta})-\nabla^2 L_i(\boldsymbol{\theta}')\|\nonumber\\
         \leq& H(1-\alpha\mu)\|\boldsymbol{\phi}_i-\boldsymbol{\phi}'_i\|+\alpha\rho B\|\boldsymbol{\theta}-\boldsymbol{\theta}'\|\\
         \leq& H'\|\boldsymbol{\theta}-\boldsymbol{\theta}'\|,
     \end{align*}
where $H'=H(1-\alpha\mu)^2+\alpha\rho B$, thereby
 completing the proof of Lemma 1.
 
\section{Proof of Theorem 1}
Observe that $\nabla G_i(\boldsymbol{\theta})=\nabla L_i(\boldsymbol{\phi}_i)-\alpha \nabla^2 L_i(\boldsymbol{\theta})\nabla L_i(\boldsymbol{\phi}_i)$  involves the product between Hessian matrix and gradient, which admits an upper bound outlined as follows: 
 \begin{small}
     \begin{align}
         &\|\nabla^2 L_i(\boldsymbol{\theta})\nabla L_i(\boldsymbol{\theta})-\sum_{i\in\mathcal{S}}\omega_i\nabla^2 L_i(\boldsymbol{\theta})\nabla L_i(\boldsymbol{\theta})\|\nonumber\\
         =&\|\nabla^2 L_i(\boldsymbol{\theta})\nabla L_i(\boldsymbol{\theta})-\nabla^2 L_w(\boldsymbol{\theta})\nabla L_w(\boldsymbol{\theta})\nonumber\\
         &+\nabla^2 L_w(\boldsymbol{\theta})\nabla L_w(\boldsymbol{\theta})-\sum_{i\in\mathcal{S}}\omega_i\nabla^2 L_i(\boldsymbol{\theta})\nabla L_i(\boldsymbol{\theta})\|\nonumber\\
         \leq& \|\nabla^2 L_i(\boldsymbol{\theta})\nabla L_i(\boldsymbol{\theta})
         -\nabla^2 L_i(\boldsymbol{\theta})\nabla L_w(\boldsymbol{\theta})\|\nonumber\\
         &+\|\nabla^2 L_i(\boldsymbol{\theta})\nabla L_w(\boldsymbol{\theta})-\nabla^2 L_w(\boldsymbol{\theta})\nabla L_w(\boldsymbol{\theta}\|\nonumber\\
         &+\|\sum_{i\in\mathcal{S}} \omega_i[(\nabla L_i(\boldsymbol{\theta})-\nabla L_w(\boldsymbol{\theta}))(\nabla^2 L_i(\boldsymbol{\theta})-\nabla^2 L_w(\boldsymbol{\theta}))]\|\nonumber\\
         \leq& \|\nabla^2 L_i(\boldsymbol{\theta})\|\|\nabla L_i(\boldsymbol{\theta})-\nabla L_w(\boldsymbol{\theta})\|\nonumber\\
         &+\|\nabla L_w(\boldsymbol{\theta})\|\|\nabla^2 L_i(\boldsymbol{\theta})-\nabla^2 L_w(\boldsymbol{\theta})\|\nonumber\\
         &+\sum_{i\in\mathcal{S}} \omega_i\|\nabla L_i(\boldsymbol{\theta})-\nabla L_w(\boldsymbol{\theta})\|\|\nabla^2 L_i(\boldsymbol{\theta})-\nabla^2 L_w(\boldsymbol{\theta})\|\nonumber\\
         \leq& H\delta_i+B\sigma_i+\tau.
     \end{align}
     \end{small}
Next, it follows from Taylor's Theorem that
     \begin{equation*}
         \nabla L_i(\boldsymbol{\phi}_i)=\nabla L_i(\boldsymbol{\theta})+\nabla^2 L_i(\boldsymbol{\theta})(\boldsymbol{\phi}_i-\boldsymbol{\theta})+O(\|\boldsymbol{\phi}_i-\boldsymbol{\theta}\|^2).
     \end{equation*}
That is to say, 
     \begin{equation*}
         \nabla L_i(\boldsymbol{\phi}_i)=\nabla L_i(\boldsymbol{\theta})-\alpha\nabla^2 L_i(\boldsymbol{\theta})\nabla L_i(\boldsymbol{\theta})+O(\alpha^2B^2).
     \end{equation*}
Therefore, we have
     \begin{small}
     \begin{align*}
         &\|\nabla G_i(\boldsymbol{\theta})-\nabla G(\boldsymbol{\theta})\|\nonumber\\
         =&\|[I-\alpha \nabla^2 L_i(\boldsymbol{\theta})]\nabla L_i(\boldsymbol{\phi}_i)
         -\sum_{i\in\mathcal{S}} \omega_i[I-\alpha \nabla^2 L_i(\boldsymbol{\theta})]\nabla L_i(\boldsymbol{\phi}_i)\|\nonumber\\
         =&\|[I-\alpha \nabla^2 L_i(\boldsymbol{\theta})][\nabla L_i(\boldsymbol{\phi}_i)-\nabla L_i(\boldsymbol{\theta})+\nabla L_i(\boldsymbol{\theta})]\nonumber\\
         &-\sum_{i\in\mathcal{S}} \omega_i[I-\alpha \nabla^2 L_i(\boldsymbol{\theta})][\nabla L_i(\boldsymbol{\phi}_i)\nabla -L_i(\boldsymbol{\theta})+\nabla L_i(\boldsymbol{\theta})]\|\nonumber\\
         =&\|\nabla L_i(\boldsymbol{\theta})-\nabla L_w(\boldsymbol{\theta})-2\alpha\nabla^2 L_i(\boldsymbol{\theta})\nabla L_i(\boldsymbol{\theta})\nonumber\\
         &+2\alpha\sum_{i\in\mathcal{S}}\omega_i\nabla^2 L_i(\boldsymbol{\theta})\nabla L_i(\boldsymbol{\theta})+O(\alpha^2B^2)\nonumber\\
         &+\alpha^2[\nabla^2 L_i(\boldsymbol{\theta})]^2\nabla L_i(\boldsymbol{\theta})-\alpha^2\sum_{i\in\mathcal{S}}\omega_i[\nabla^2 L_i(\boldsymbol{\theta})]^2\nabla L_i(\boldsymbol{\theta})\|\nonumber\\
         \leq&\|\nabla L_i(\boldsymbol{\theta})-\nabla L_w(\boldsymbol{\theta})\|
         +O(\alpha^2B^2)\nonumber\\
         &+2\alpha\|\nabla^2 L_i(\boldsymbol{\theta})\nabla L_i(\boldsymbol{\theta})-\sum_{i\in\mathcal{S}}\omega_i\nabla^2 L_i(\boldsymbol{\theta})\nabla L_i(\boldsymbol{\theta})\|\nonumber\\
         &+\alpha^2\|[\nabla^2 L_i(\boldsymbol{\theta})]^2\nabla L_i(\boldsymbol{\theta})-\sum_{i\in\mathcal{S}}\omega_i[\nabla^2 L_i(\boldsymbol{\theta})]^2\nabla L_i(\boldsymbol{\theta})\|\nonumber\\
         \leq& \delta_i+2\alpha(H\delta_i+B\sigma_i+\tau)+O(\alpha^2B^2)\nonumber\\
         &+\alpha^2\|[\nabla^2 L_i(\boldsymbol{\theta})]^2\nabla L_i(\boldsymbol{\theta})-\alpha^2\sum_{i\in\mathcal{S}}\omega_i[\nabla^2 L_i(\boldsymbol{\theta})]^2\nabla L_i(\boldsymbol{\theta})\|.
     \end{align*}
     \end{small}
Along the same line as in finding an upper bound for $\|\nabla^2 L_i(\boldsymbol{\theta})\nabla L_i(\boldsymbol{\theta})-\sum_{i\in\mathcal{S}}\omega_i\nabla^2 L_i(\boldsymbol{\theta})\nabla L_i(\boldsymbol{\theta})\|$, we can find an upper bound on  the last term of the above inequality with $\alpha^2(H\delta'_i+B\sigma_i+\tau')$ where $\delta'_i=H\delta_i+B\sigma_i+\tau$ and $\tau'=\sum_{i\in\mathcal{S}} \omega_i\delta'_i\sigma_i$. We conclude that when $\alpha$ is suitably small, there exists a constant $C$ such that the right hand side of the above inequality is upper bounded by $\delta_i+\alpha C(H\delta_i+B\sigma_i+\tau)$.
 
\section{Proof of Theorem 2}
Following the same method as in \cite{wang2019}, we first define a virtual sequence for global aggregation at each iteration as $\boldsymbol{v}_{[n]}^t$ for $t\in[(n-1)T_0, nT_0]$, where the interval $[(n-1)T_0, nT_0]$ is denoted as $[n]$. More specifically,
\begin{equation}\label{virtual}
       \boldsymbol{v}_{[n]}^{t+1}=\boldsymbol{v}_{[n]}^t-\beta\nabla G(\boldsymbol{v}_{[n]}^t),
\end{equation}
and $\boldsymbol{v}_{[n]}^{t}$ is assumed to be ``synchronized" with $\boldsymbol{\theta}^t$ at the beginning of interval $[n]$, i.e., $\boldsymbol{v}_{[n]}^{(n-1)T_0}=\boldsymbol{\theta}^{(n-1)T_0}$, where $\boldsymbol{\theta}^{(n-1)T_0}$ is the weighted average of local parameters $\boldsymbol{\theta}_i^{(n-1)T_0}$ as shown in \eqref{aggregation}. To show the convergence, we first analyze the gap between the virtual global  parameter $\boldsymbol{v}_{[n]}^{t}$ and the local weighted average $\boldsymbol{\theta}^{t}$ during each interval, and then evaluate the convergence performance of $\boldsymbol{\theta}^{t}$ through evaluating the convergence performance of virtual sequence $\boldsymbol{v}_{[n]}^{t}$ by taking the gap into consideration.

To analyze the gap between $\boldsymbol{v}_{[n]}^{t}$ and $\boldsymbol{\theta}^{t}$ during interval $[n]$, we first look into the gap between $\boldsymbol{v}_{[n]}^{t}$ and the local update $\boldsymbol{\theta}_i^{t}$. Specifically, 
     \begin{small}
     \begin{align}\label{theta_i}
         &\|\boldsymbol{\theta}_i^{t+1}-\boldsymbol{v}_{[n]}^{t+1}\|\nonumber\\
         =&\|\boldsymbol{\theta}_i^{t}-\beta\nabla G_i(\boldsymbol{\theta}_i^{t})-\boldsymbol{v}_{[n]}^{t}+\beta\nabla G(\boldsymbol{v}_{[n]}^{t})\|\nonumber\\
         \leq&\|\boldsymbol{\theta}_i^{t}-\boldsymbol{v}_{[n]}^{t}\|+\beta\|\nabla G_i(\boldsymbol{\theta}_i^{t})-\nabla G(\boldsymbol{v}_{[n]}^{t})\|\nonumber\\
         \leq&\|\boldsymbol{\theta}_i^{t}-\boldsymbol{v}_{[n]}^{t}\|+\beta\|\nabla G_i(\boldsymbol{\theta}_i^{t})-\nabla G_i(\boldsymbol{v}_{[n]}^{t})\|\nonumber\\
         &+\beta\|\nabla G_i(\boldsymbol{v}_{[n]}^{t})-\nabla G(\boldsymbol{v}_{[n]}^{t})\|\nonumber\\
         \leq&(1+\beta H')\|\boldsymbol{\theta}_i^{t}-\boldsymbol{v}_{[n]}^{t}\|+\beta[\delta_i+\alpha C(H\delta_i+B\sigma_i+\tau)].
     \end{align}
     \end{small}
By induction, we can show that $\|\boldsymbol{\theta}_i^{t}-\boldsymbol{v}_{[n]}^{t}\|\leq g(t-(n-1)T_0)$ where $g(x)\triangleq \frac{\delta_i+\alpha C(H\delta_i+B\sigma_i+\tau)}{H'}[(1+\beta H')^{x}-1]$. Therefore, for $t\in [(n-1)T_0, nT_0)$ we can get
     \begin{small}
         \begin{align}\label{theta}
             &\|\boldsymbol{\theta}^{t+1}-\boldsymbol{v}_{[n]}^{t+1}\|\nonumber\\
             =&\|\sum_{i\in\mathcal{S}}\omega_i\boldsymbol{\theta}_i^{t+1}-\boldsymbol{v}_{[n]}^{t+1}\|\nonumber\\
             =&\|\boldsymbol{\theta}^{t}-\beta\sum_{i\in\mathcal{S}}\omega_i\nabla G_i(\boldsymbol{\theta}_i^t)-\boldsymbol{v}_{[n]}^{t}+\beta\nabla G(\boldsymbol{v}_{[n]}^{t})\|\nonumber\\
             \leq&\|\boldsymbol{\theta}^{t}-\boldsymbol{v}_{[n]}^{t}\|+\beta\|\sum_{i\in\mathcal{S}}\omega_i(\nabla G_i(\boldsymbol{\theta}_i^t)-\nabla G_i(\boldsymbol{v}_{[n]}^{t}))\|\nonumber\\
             \leq&\|\boldsymbol{\theta}^{t}-\boldsymbol{v}_{[n]}^{t}\|+\beta\sum_{i\in\mathcal{S}}\omega_i\|\nabla G_i(\boldsymbol{\theta}_i^t)-\nabla G_i(\boldsymbol{v}_{[n]}^{t})\|\nonumber\\
             \leq&\|\boldsymbol{\theta}^{t}-\boldsymbol{v}_{[n]}^{t}\|+\beta H'\sum_{i\in\mathcal{S}}\omega_i\|\boldsymbol{\theta}_i^t-\boldsymbol{v}_{[n]}^{t}\|\nonumber\\
             \leq&\|\boldsymbol{\theta}^{t}-\boldsymbol{v}_{[n]}^{t}\|+\beta H'\sum_{i\in\mathcal{S}}\omega_i g(t-(n-1)T_0)\nonumber\\
             =&\|\boldsymbol{\theta}^{t}-\boldsymbol{v}_{[n]}^{t}\|+\alpha'[(1+\beta H')^{t-(n-1)T_0}-1],
         \end{align}
     \end{small}
where $\alpha'=\beta[\delta+\alpha C(H\delta+B\sigma+\tau)]$. Since $\boldsymbol{\theta}^{(n-1)T_0}=\boldsymbol{v}_{[n]}^{(n-1)T_0}$, we have
     \begin{small}
         \begin{align}\label{h()}
             \|\boldsymbol{\theta}^{t}-\boldsymbol{v}_{[n]}^{t}\|
             \leq& \sum_{j=1}^{t-(n-1)T_0}\{\alpha'[(1+\beta H')^j-1]\}\nonumber\\
             =&\frac{\alpha'}{\beta H'}[(1+\beta H')^{t-(n-1)T_0}-1]-\alpha'[t-(n-1)T_0]\nonumber\\
             \triangleq&h(t-(n-1)T_0).
         \end{align}
     \end{small}
 Next, we evaluate the convergence performance of virtual sequence $\boldsymbol{v}_{[n]}^t$ during the interval $[n]$ for $t\in[(n-1)T_0,nT_0]$. Since $G(\cdot)$ is $H'$-smooth, we can have
 \begin{small}
 \begin{align}\label{G-lip}
     G(\boldsymbol{v}_{[n]}^{t+1})-G(\boldsymbol{v}_{[n]}^t) \leq& \langle\nabla G(\boldsymbol{v}_{[n]}^t), \boldsymbol{v}_{[n]}^{t+1}-\boldsymbol{v}_{[n]}^t\rangle\nonumber\\
     &+\frac{H'}{2}\|\boldsymbol{v}_{[n]}^{t+1}-\boldsymbol{v}_{[n]}^t\|^2\nonumber\\
     \leq& -\beta\left(1-\frac{H'\beta}{2}\right)\|\nabla G(\boldsymbol{v}_{[n]}^t)\|^2.
 \end{align}
 \end{small}
 Moreover, since $G(\cdot)$ is $\mu'$-strongly convex, it follows that
 \begin{equation}\label{G-convex}
     G(\boldsymbol{v}_{[n]}^t)\leq G(\boldsymbol{\theta}^\star)+\frac{1}{2\mu'}\|\nabla G(\boldsymbol{v}_{[n]}^t)\|^2.
 \end{equation}
 Combining \eqref{G-lip} and \eqref{G-convex} gives us 
 \begin{small}
 \begin{equation*}
     G(\boldsymbol{v}_{[n]}^{t+1})-G(\boldsymbol{v}_{[n]}^t) \leq -2\beta\mu'\left(1-\frac{H'\beta}{2}\right)[G(\boldsymbol{v}_{[n]}^t)- G(\boldsymbol{\theta}^\star)]
 \end{equation*}
 \end{small}
 which is equivalent with
  \begin{small}
 \begin{align*}
     G(\boldsymbol{v}_{[n]}^{t+1})-G(\boldsymbol{\theta}^\star) &\leq \left[1-2\beta\mu'\left(1-\frac{H'\beta}{2}\right)\right][G(\boldsymbol{v}_{[n]}^t)- G(\boldsymbol{\theta}^\star)]\\
     &=\xi[G(\boldsymbol{v}_{[n]}^t)- G(\boldsymbol{\theta}^\star)],
 \end{align*}
 \end{small}
 where $\xi=1-2\beta\mu'\left(1-\frac{H'\beta}{2}\right)\in(0,1)$ given $\beta<\min\{\frac{1}{2\mu'},\frac{2}{H'}\}$. Iteratively, we can obtain
 \begin{small}
 \begin{align}\label{intervalconv}
     G(\boldsymbol{v}_{[n]}^{nT_0})-G(\boldsymbol{\theta}^\star)
     \leq& \xi[G(\boldsymbol{v}_{[n]}^{nT_0-1})-G(\boldsymbol{\theta}^\star)]\nonumber\\
     \leq & \xi^2[G(\boldsymbol{v}_{[n]}^{nT_0-2})-G(\boldsymbol{\theta}^\star)]\nonumber\\
     ...&\nonumber\\
     \leq& \xi^{T_0}[G(\boldsymbol{v}_{[n]}^{(n-1)T_0})-G(\boldsymbol{\theta}^\star)]\nonumber\\
     =&\xi^{T_0}[G(\boldsymbol{v}_{[n-1]}^{(n-1)T_0})-G(\boldsymbol{\theta}^\star)]\nonumber\\
     &+\xi^{T_0}[G(\boldsymbol{v}_{[n]}^{(n-1)T_0})-G(\boldsymbol{v}_{[n-1]}^{(n-1)T_0})].
 \end{align}
 \end{small}
 Note that 
 \begin{small}
         \begin{align}\label{upperG}
             \|\nabla G(\boldsymbol{\theta})\|&=\|\sum_{i\in\mathcal{S}}\omega_i[(I-\alpha\nabla^2 L_i(\boldsymbol{\theta}))\nabla L_i(\boldsymbol{\phi}_i)]\|\nonumber\\
             &\leq\sum_{i\in\mathcal{S}}\omega_i\|I-\alpha\nabla^2 L_i(\boldsymbol{\theta})\|\|\nabla L_i(\boldsymbol{\phi}_i)\|\nonumber\\
             &\leq(1-\alpha\mu)B.
        \end{align}
     \end{small}
From the Mean Value Theorem, we conclude that $\|G(\boldsymbol{\theta})-G(\boldsymbol{\theta}')\|\leq (1-\alpha\mu)B\|\boldsymbol{\theta}-\boldsymbol{\theta}'\|$. Hence, we can upper bound $G(\boldsymbol{v}_{[n]}^{(n-1)T_0})-G(\boldsymbol{v}_{[n-1]}^{(n-1)T_0})$ as follows.
\begin{small}
\begin{align}\label{endbound}
    &G(\boldsymbol{v}_{[n]}^{(n-1)T_0})-G(\boldsymbol{v}_{[n-1]}^{(n-1)T_0})\nonumber\\
    =&G(\boldsymbol{\theta}^{(n-1)T_0})-G(\boldsymbol{v}_{[n-1]}^{(n-1)T_0})\nonumber\\
    \leq& B(1-\alpha\mu)\|\boldsymbol{\theta}^{(n-1)T_0}-\boldsymbol{v}_{[n-1]}^{(n-1)T_0}\|\nonumber\\
    \overset{(a)}{\leq}& B(1-\alpha\mu)h(T_0),
\end{align}
\end{small}
where (a) is from \eqref{h()}. Substitute \eqref{endbound} in \eqref{intervalconv}, we have
\begin{small}
\begin{align*}
    &G(\boldsymbol{v}_{[n]}^{nT_0})-G(\boldsymbol{\theta}^\star)\\
    \leq &\xi^{T_0}[G(\boldsymbol{v}_{[n-1]}^{(n-1)T_0})-G(\boldsymbol{\theta}^\star)]+\xi^{T_0}B(1-\alpha\mu)h(T_0).
\end{align*}
\end{small}
Iteratively, it follows that
\begin{small}
\begin{align*}
    &G(\boldsymbol{v}_{[N]}^{NT_0})-G(\boldsymbol{\theta}^\star)\\
    \leq &\xi^{T_0}[G(\boldsymbol{v}_{[N-1]}^{(N-1)T_0})-G(\boldsymbol{\theta}^\star)]+\xi^{T_0}B(1-\alpha\mu)h(T_0)\\
    \leq & \xi^{2T_0}[G(\boldsymbol{v}_{[N-2]}^{(N-2)T_0})-G(\boldsymbol{\theta}^\star)]+(\xi^{T_0}+\xi^{2T_0})B(1-\alpha\mu)h(T_0)\\
    ... &\\
    \leq & \xi^{(N-1)T_0}[G(\boldsymbol{v}_{[1]}^{T_0})-G(\boldsymbol{\theta}^\star)]+\sum_{j=1}^{N-1} \xi^{jT_0}B(1-\alpha\mu)h(T_0) \\
    \leq & \xi^{NT_0}[G(\boldsymbol{v}_{[1]}^{0})-G(\boldsymbol{\theta}^\star)]+\sum_{j=1}^{N-1} \xi^{jT_0}B(1-\alpha\mu)h(T_0) \\
    =&\xi^{NT_0}[G(\boldsymbol{\theta}^0)-G(\boldsymbol{\theta}^\star)]+\sum_{j=1}^{N-1} \xi^{jT_0}B(1-\alpha\mu)h(T_0). 
\end{align*}
\end{small}
Therefore, we can conclude that
\begin{small}
\begin{align*}
    &G(\boldsymbol{\theta}^T)-G(\boldsymbol{\theta}^\star)\\
    =&G(\boldsymbol{v}_{[N+1]}^{NT_0})-G(\boldsymbol{\theta}^\star)\\
    =&G(\boldsymbol{v}_{[N]}^{NT_0})-G(\boldsymbol{\theta}^\star)+G(\boldsymbol{v}_{[N+1]}^{NT_0})-G(\boldsymbol{v}_{[N]}^{NT_0})\\
    \leq&\xi^{NT_0}[G(\boldsymbol{\theta}^0)-G(\boldsymbol{\theta}^\star)]+\sum_{j=1}^{N-1} \xi^{jT_0}B(1-\alpha\mu)h(T_0)\\
    &+B(1-\alpha\mu)h(T_0)\\
    =&\xi^T[G(\boldsymbol{\theta}^0)-G(\boldsymbol{\theta}^\star)]+\frac{1-\xi^T}{1-\xi^{T_0}}B(1-\alpha\mu)h(T_0)\\
    \leq &\xi^T[G(\boldsymbol{\theta}^0)-G(\boldsymbol{\theta}^\star)]+\frac{B(1-\alpha\mu)}{1-\xi^{T_0}}h(T_0),
\end{align*}
\end{small}
thereby completing the proof of Theorem 2.

\section{Proof of Theorem 3}
Recall that  the optimal model parameter is denoted as $\boldsymbol{\phi}^\star_t=\boldsymbol{\theta}_t^\star-\alpha\nabla L^\star_t(\boldsymbol{\theta}_t^\star)$, i.e., $\boldsymbol{\phi}^\star_t$ can be obtained through one-step gradient update from parameter $\boldsymbol{\theta}_t^\star$. However, through the fast adaptation from the meta-learned model $\boldsymbol{\theta}_c$, we have $\boldsymbol{\phi_t}=\boldsymbol{\theta}_c-\alpha\nabla L_t(\boldsymbol{\theta}_c)$, where $\boldsymbol{\theta}_c$ can be regarded as an estimation of $\boldsymbol{\theta}_t^\star$ and $L_t(\cdot)$ is the sample average approximation of $L^\star_t(\cdot)$. To evaluate the learning performance at the target, we next evaluate the gap between $L^\star_t(\boldsymbol{\phi}^\star_t)$ and $L^\star_t(\boldsymbol{\phi}_t)$, i.e., the gap between the optimal loss and the actual loss based on learned parameter $\boldsymbol{\phi}_t$.

For convenience, denote $\Tilde{\boldsymbol{\phi}}_t=\boldsymbol{\theta}_t^\star-\alpha\nabla L_t(\boldsymbol{\theta}_t^\star)$.  Note that
     \begin{small}
     \begin{align}
         \|\boldsymbol{\phi}_t-\boldsymbol{\phi}^\star_t\|&=\|\boldsymbol{\phi}_t-\Tilde{\boldsymbol{\phi}}_t+\Tilde{\boldsymbol{\phi}}_t-\boldsymbol{\phi}^\star_t\|\nonumber\\
         &\leq \underbrace{\|\boldsymbol{\phi}_t-\Tilde{\boldsymbol{\phi}}_t\|}_{(a)}+\underbrace{\|\Tilde{\boldsymbol{\phi}}_t-\boldsymbol{\phi}^\star_t\|}_{(b)},
     \end{align}
     \end{small}
where (a) represents the error introduced by the gap between meta-learned model $\boldsymbol{\theta}_c$ and the target optimal $\boldsymbol{\theta}_t^\star$, and (b) captures the error from the sample average approximation of the loss function.
    
We have the following bound on the term in (a): 
     \begin{small}
         \begin{align}\label{modeldiff}
              \|\boldsymbol{\phi}_t-\Tilde{\boldsymbol{\phi}}_t\|&=\|\boldsymbol{\theta}_c-\boldsymbol{\theta}_t^\star-\alpha(\nabla L_t(\boldsymbol{\theta}_c)-\nabla L_t(\boldsymbol{\theta}_t^\star))\|\nonumber\\
     &\leq \|\boldsymbol{\theta}_c-\boldsymbol{\theta}_t^\star\|+\alpha\|\nabla L_t(\boldsymbol{\theta}_c)-\nabla L_t(\boldsymbol{\theta}_t^\star)\|\nonumber\\
     &\leq (1+\alpha H)\|\boldsymbol{\theta}_c-\boldsymbol{\theta}_t^\star\|\nonumber\\
     &=(1+\alpha H)\|\boldsymbol{\theta}_c-\boldsymbol{\theta}_c^\star+\boldsymbol{\theta}_c^\star-\boldsymbol{\theta}_t^\star\|\nonumber\\
     &\leq (1+\alpha H)[\|\boldsymbol{\theta}_c-\boldsymbol{\theta}_c^\star\|+\|\boldsymbol{\theta}_c^\star-\boldsymbol{\theta}_t^\star\|]\nonumber\\
     &\leq (1+\alpha 
     H)[\epsilon_c+\|\boldsymbol{\theta}_c^\star-\boldsymbol{\theta}_t^\star\|].
         \end{align}
     \end{small}
     
To evaluate the term in (b), we first note that 
$\|\Tilde{\boldsymbol{\phi}}_t-\boldsymbol{\phi}^\star_t\|=\alpha\|\nabla L_t(\boldsymbol{\theta}_t^\star)-\nabla L^\star_t(\boldsymbol{\theta}_t^\star)\|$. Here, $\nabla L_t(\cdot)=\frac{1}{K}\sum_{(\mathbf{x}_t^j,\mathbf{y}_t^j)\in D_t}\nabla l(\cdot, (\mathbf{x}_t^j,\mathbf{y}_t^j))$, and $\nabla L^\star_t(\cdot)=\mathbb{E}_{(\mathbf{x}_t,\mathbf{y}_t)\sim P_t}\nabla l(\cdot, (\mathbf{x}_t,\mathbf{y}_t))$. Define $q_t(\cdot)\triangleq\nabla l_t(\cdot)$. Then
 \begin{equation*}
     Q_t(\boldsymbol{\theta}_t^\star)\triangleq\frac{1}{K}\sum_{(\mathbf{x}_t^j,\mathbf{y}_t^j)\in D_t}q_t(\boldsymbol{\theta}_t^\star, (\mathbf{x}_t^j,\mathbf{y}_t^j))=\nabla L_t(\boldsymbol{\theta}_t^\star),
 \end{equation*}
and
 \begin{equation*}
     Q^\star_t(\boldsymbol{\theta}_t^\star)\triangleq\mathbb{E}_{(\mathbf{x}_t,\mathbf{y}_t)\sim P_t}q_t(\boldsymbol{\theta}_t^\star, (\mathbf{x}_t,\mathbf{y}_t))=\nabla L^\star_t(\boldsymbol{\theta}_t^\star).
 \end{equation*}
Clearly $Q_t(\cdot)$ is the sample average approximation of $Q^\star_t(\cdot)$. Since $l(\boldsymbol{\theta},(\mathbf{x}_t^j, \mathbf{y}_t^j))$ is $H$-smooth for all $\boldsymbol{\theta}\in\mathbb{R}^d$ and $(\mathbf{x}_t^j, \mathbf{y}_t^j)\sim P_t$, from Theorem 7.73 in \cite{Shapiro2009} about uniform law of large number, we can know that for any $\epsilon>0$ there exist positive constants $C_t$ and $\eta=\eta(\epsilon)$ such that
 \begin{small}
 \begin{equation*}
     Pr\left\{\sup_{\boldsymbol{\theta}\in\Theta}\|Q_t(\boldsymbol{\theta})-Q^\star_t(\boldsymbol{\theta})\|\geq \epsilon\right\}\leq C_te^{-K\eta},
 \end{equation*}
 \end{small}
where $\Theta=\{\boldsymbol{\theta}|\|\boldsymbol{\theta}-\boldsymbol{\theta}_c\|\leq D\}$ and $D$ is some constant. Hence, we have
 \begin{small}
 \begin{equation}\label{sample}
     Pr\left\{\|\Tilde{\boldsymbol{\phi}}_t-\boldsymbol{\phi}^\star_t\|\leq \alpha\epsilon\right\}\geq 1-C_te^{-K\eta}.
 \end{equation}
 \end{small}
Since $L^\star_t(\cdot)$ is also $H$-smooth, combing \eqref{modeldiff} and \eqref{sample} proves Theorem 3.
 
\section{Proof of Theorem 4}
Let $\Tilde{l}(\boldsymbol{\theta},\mathbf{x})\triangleq l(\boldsymbol{\theta},(\mathbf{x},\mathbf{y}_0))-\lambda c((\mathbf{x},\mathbf{y}_0),(\mathbf{x}_0,\mathbf{y}_0))$, then the robust surrogate loss $l_\lambda(\boldsymbol{\theta},(\mathbf{x}_0,\mathbf{y}_0))=\sup_{\mathbf{x}\in\mathcal{X}}\Tilde{l}(\boldsymbol{\theta},\mathbf{x})$. Based on Assumption 5 and 7, it can be easily shown that $\Tilde{l}(\boldsymbol{\theta},\mathbf{x})$ is $(\lambda-H_{\mathbf{xx}})$-strongly concave with respect to $\mathbf{x}$. 
Let $\mathbf{x}^{\star}(\boldsymbol{\theta})=argmax_{\mathbf{x}\in\mathcal{X}}\Tilde{l}(\boldsymbol{\theta},\mathbf{x})$. From Lemma 1 in \cite{sinha2017certifying}, we conclude that $l_{\lambda}$ is differentiable and $\nabla_{\boldsymbol{\theta}}l_\lambda(\boldsymbol{\theta},(\mathbf{x}_0,\mathbf{y}_0))=\nabla_{\boldsymbol{\theta}}\Tilde{l}(\boldsymbol{\theta},\mathbf{x}^\star(\boldsymbol{\theta}))$. Moreover, 
\begin{equation}
    \|\mathbf{x}^\star(\boldsymbol{\theta})-\mathbf{x}^\star(\boldsymbol{\theta}')\|\leq \frac{H_{\mathbf{x}\boldsymbol{\theta}}}{\lambda-H_{\mathbf{xx}}}\|\boldsymbol{\theta}-\boldsymbol{\theta}'\|,
\end{equation}
and
\begin{align*}
    \|\nabla_{\boldsymbol{\theta}}l_\lambda(\boldsymbol{\theta})-\nabla_{\boldsymbol{\theta}}l_\lambda(\boldsymbol{\theta}')\|\leq& \left(H+\frac{H_{\boldsymbol{\theta}\mathbf{x}}H_{\mathbf{x}\boldsymbol{\theta}}}{\lambda-H_{\mathbf{xx}}}\right)\|\boldsymbol{\theta}-\boldsymbol{\theta}'\|\\
    \triangleq& H_{\lambda}\|\boldsymbol{\theta}-\boldsymbol{\theta}'\|.
\end{align*}
Besides,
\begin{align*}
    &\|\nabla_{\boldsymbol{\theta}}l_\lambda(\boldsymbol{\theta})-\nabla_{\boldsymbol{\theta}}l_\lambda(\boldsymbol{\theta}')\|\\
    =&\|\nabla_{\boldsymbol{\theta}}\Tilde{l}(\boldsymbol{\theta},\mathbf{x}^\star(\boldsymbol{\theta}))-\nabla_{\boldsymbol{\theta}}\Tilde{l}(\boldsymbol{\theta}',\mathbf{x}^\star(\boldsymbol{\theta}'))\|\\
    =&\|\nabla_{\boldsymbol{\theta}}\Tilde{l}(\boldsymbol{\theta},\mathbf{x}^\star(\boldsymbol{\theta}))-\nabla_{\boldsymbol{\theta}}\Tilde{l}(\boldsymbol{\theta},\mathbf{x}^\star(\boldsymbol{\theta}'))
    +\nabla_{\boldsymbol{\theta}}\Tilde{l}(\boldsymbol{\theta},\mathbf{x}^\star(\boldsymbol{\theta}'))\\
    &-\nabla_{\boldsymbol{\theta}}\Tilde{l}(\boldsymbol{\theta}',\mathbf{x}^\star(\boldsymbol{\theta}'))\|\\
    \geq&\|\nabla_{\boldsymbol{\theta}}\Tilde{l}(\boldsymbol{\theta},\mathbf{x}^\star(\boldsymbol{\theta}'))-\nabla_{\boldsymbol{\theta}}\Tilde{l}(\boldsymbol{\theta}',\mathbf{x}^\star(\boldsymbol{\theta}'))\|\\
    &-\|\nabla_{\boldsymbol{\theta}}\Tilde{l}(\boldsymbol{\theta},\mathbf{x}^\star(\boldsymbol{\theta}))-\nabla_{\boldsymbol{\theta}}\Tilde{l}(\boldsymbol{\theta},\mathbf{x}^\star(\boldsymbol{\theta}'))\|\\
    \geq& \mu\|\boldsymbol{\theta}-\boldsymbol{\theta}'\|-H_{\boldsymbol{\theta}\mathbf{x}}\|\mathbf{x}^\star(\boldsymbol{\theta})-\mathbf{x}^\star(\boldsymbol{\theta}')\|\\
    \geq& \left(\mu-\frac{H_{\boldsymbol{\theta}\mathbf{x}}H_{\mathbf{x}\boldsymbol{\theta}}}{\lambda-H_{\mathbf{xx}}}\right)\|\boldsymbol{\theta}-\boldsymbol{\theta}'\|\triangleq \mu_{\lambda}\|\boldsymbol{\theta}-\boldsymbol{\theta}'\|.
\end{align*}
So the strongly convexity and smoothness of the robust surrogate loss $l_{\lambda}$ still hold if $\lambda$ is large enough. Based on the triangle inequality, $\mathbb{E}_{P_i}[l_{\lambda}(\boldsymbol{\theta},(\mathbf{x}_i,\mathbf{y}_i))]$ is $\mu_{\lambda}$-strongly convex and $H_{\lambda}$-smooth both with respect to $\boldsymbol{\theta}$ for all $i\in\mathcal{S}$.

Denote $\Tilde{H}_i(\boldsymbol{\theta})=\mathbb{E}_{P_i}[l_{\lambda}(\boldsymbol{\theta},(\mathbf{x}_i,\mathbf{y}_i))]$. Then we have
\begin{small}
\begin{align}\label{robustconvex}
    &\|\nabla_{\boldsymbol{\theta}} \mathbb{E}_{P_i}[l_{\lambda}(\boldsymbol{\phi}_i(\boldsymbol{\theta}),(\mathbf{x}_i,\mathbf{y}_i))]-\nabla_{\boldsymbol{\theta}}\mathbb{E}_{P_i}[l_{\lambda}(\boldsymbol{\phi}_i'(\boldsymbol{\theta}'),(\mathbf{x}_i,\mathbf{y}_i))]\|\nonumber\\
    =&\|\nabla \Tilde{H}_i(\boldsymbol{\phi}_i)[I-\alpha\nabla^2 L_i(\boldsymbol{\theta})]-\nabla\Tilde{H}_i(\boldsymbol{\phi}_i')[I-\alpha\nabla^2 L_i(\boldsymbol{\theta}')]\|\nonumber\\
    =&\|\nabla \Tilde{H}_i(\boldsymbol{\phi}_i)[I-\alpha\nabla^2 L_i(\boldsymbol{\theta})]-\nabla\Tilde{H}_i(\boldsymbol{\phi}_i')[I-\alpha\nabla^2 L_i(\boldsymbol{\theta}')]\nonumber\\
    &+\alpha\nabla\Tilde{H}_i(\boldsymbol{\phi}_i')\nabla^2 L_i(\boldsymbol{\theta})-\alpha\nabla\Tilde{H}_i(\boldsymbol{\phi}_i')\nabla^2 L_i(\boldsymbol{\theta})\|\nonumber\\
    =&\|[\nabla\Tilde{H}_i(\boldsymbol{\phi}_i)-\nabla\Tilde{H}_i(\boldsymbol{\phi}_i')][I-\alpha\nabla^2 L_i(\boldsymbol{\theta})]\nonumber\\
    &-\alpha\nabla\Tilde{H}_i(\boldsymbol{\phi}_i')[\nabla^2 L_i(\boldsymbol{\theta})-\nabla^2 L_i(\boldsymbol{\theta}')]\|.
\end{align}
\end{small}
Since $\nabla \boldsymbol{\phi}_i=I-\alpha\nabla^2 L_i(\boldsymbol{\theta})$, it follows from Assumption 6 and 7 that $1-\alpha H\leq \nabla \boldsymbol{\phi}_i\leq 1-\alpha\mu$, which indicates
 \begin{equation*}
     (1-\alpha H)\|\boldsymbol{\theta}-\boldsymbol{\theta}'\|\leq \|\boldsymbol{\phi}_i-\boldsymbol{\phi}'_i\|\leq (1-\alpha\mu)\|\boldsymbol{\theta}-\boldsymbol{\theta}'\|.
 \end{equation*}
Continuing with \eqref{robustconvex}, it is clear that
\begin{small}
\begin{align*}
    &\|\nabla_{\boldsymbol{\theta}} \mathbb{E}_{P_i}[l_{\lambda}(\boldsymbol{\phi}_i(\boldsymbol{\theta}),(\mathbf{x}_i,\mathbf{y}_i))]-\nabla_{\boldsymbol{\theta}}\mathbb{E}_{P_i}[l_{\lambda}(\boldsymbol{\phi}_i'(\boldsymbol{\theta}'),(\mathbf{x}_i,\mathbf{y}_i))]\|\nonumber\\
    \geq& (1-\alpha H)\|\nabla\Tilde{H}_i(\boldsymbol{\phi}_i)-\nabla\Tilde{H}_i(\boldsymbol{\phi}_i')\|\\
    &-\alpha B\|\nabla^2 L_i(\boldsymbol{\theta})-\nabla^2 L_i(\boldsymbol{\theta}')\|\\
    \geq&\mu_{\lambda}(1-\alpha H)\|\boldsymbol{\phi}_i-\boldsymbol{\phi}_i'\|-\alpha B\rho \|\boldsymbol{\theta}-\boldsymbol{\theta}'\|\\
    \geq&[\mu_{\lambda}(1-\alpha H)^2-\alpha B\rho]\|\boldsymbol{\theta}-\boldsymbol{\theta}'\|,
\end{align*}
\end{small}
and 
\begin{small}
\begin{align*}
    &\|\nabla_{\boldsymbol{\theta}} \mathbb{E}_{P_i}[l_{\lambda}(\boldsymbol{\phi}_i(\boldsymbol{\theta}),(\mathbf{x}_i,\mathbf{y}_i))]-\nabla_{\boldsymbol{\theta}}\mathbb{E}_{P_i}[l_{\lambda}(\boldsymbol{\phi}_i'(\boldsymbol{\theta}'),(\mathbf{x}_i,\mathbf{y}_i))]\|\nonumber\\
    \leq&(1-\alpha\mu)\|\nabla\Tilde{H}_i(\boldsymbol{\phi}_i)-\nabla\Tilde{H}_i(\boldsymbol{\phi}_i')\|\\
    &+\alpha B\|\nabla^2 L_i(\boldsymbol{\theta})-\nabla^2 L_i(\boldsymbol{\theta}')\|\\
    \leq& H_{\lambda}(1-\alpha\mu)\|\boldsymbol{\phi}_i-\boldsymbol{\phi}_i'\|+\alpha B\rho \|\boldsymbol{\theta}-\boldsymbol{\theta}'\|\\
    \leq&[H_{\lambda}(1-\alpha\mu)^2+\alpha B\rho]\|\boldsymbol{\theta}-\boldsymbol{\theta}'\|.
\end{align*}
\end{small}
From the definition of $\Tilde{G}_i(\boldsymbol{\theta})$, we can have that $\Tilde{G}_i(\boldsymbol{\theta})$ is $\mu_R$-strongly convex and $H_R$-smooth, where $\mu_R=\mu+\mu_{\lambda}(1-\alpha H)^2-\alpha B\rho$ and $H_R=H+H_{\lambda}(1-\alpha\mu)^2+\alpha B\rho$. Based on the triangle inequality, Theorem 4 can be proved.

\end{document}